\documentclass[letterpaper, 10 pt, conference]{ieeeconf}

\usepackage{times}
\usepackage{multicol}
\usepackage[bookmarks=true]{hyperref}

\usepackage{soul}
\usepackage{graphicx}
\usepackage{epsfig}
\usepackage{amsmath}
\usepackage{amssymb}

\usepackage{bm}
\usepackage{booktabs}
\usepackage{color}
\usepackage{balance}
\usepackage[utf8]{inputenc}

\usepackage[ruled,linesnumbered]{algorithm2e}

\usepackage{verbatim}
\usepackage{cprotect}

\usepackage{cite}
\usepackage{subcaption}

\IEEEoverridecommandlockouts
\usepackage{balance}
\usepackage{lastpage}

\usepackage{multirow}
\usepackage{makecell}

\DeclareGraphicsExtensions{.jpg,.pdf,.png,.eps}
\graphicspath{{images/} } 

\begin{document}

\title{\LARGE \bf $\tt{ROBUSfT}$: Robust Real-Time Shape-from-Template, a C++ Library}

\author{Mohammadreza Shetab-Bushehri, Miguel Aranda, Youcef Mezouar, Adrien Bartoli, Erol \"{O}zg\"{u}r\\
\thanks{MR. Shetab-Bushehri, Y. Mezouar, A. Bartoli, and E. \"{O}zg\"{u}r are with the CNRS, Clermont Auvergne INP, Institut Pascal, Universit\'{e} Clermont Auvergne, F-63000 Clermont-Ferrand, France (e-mail: m.r.shetab@gmail.com; \hspace{3.5pt}  youcef.mezouar@sigma-clermont.fr; adrien.bartoli@gmail.com; erolozgur@gmail.com).
M. Aranda is with Instituto de Investigación en Ingeniería de Aragón (I3A), Universidad de Zaragoza, E-50018 Zaragoza, Spain (e-mail: miguel.aranda@unizar.es). }
}
\color{black}

\maketitle
\begin{abstract}
Tracking the 3D shape of a deforming object using only monocular 2D vision is a challenging problem.
This is because one should \textit{(i)} infer the 3D shape from a 2D image, which is a severely underconstrained problem, and \textit{(ii)} implement the whole solution pipeline in real-time.
The pipeline typically requires feature detection and matching, mismatch filtering, 3D shape inference and feature tracking algorithms. We propose $\tt{ROBUSfT}$, a conventional pipeline based on a template containing the object's rest shape, texturemap and deformation law. 
$\tt{ROBUSfT}$ is ready-to-use, wide-baseline, capable of handling large deformations, fast up to 30\,fps, free of training, and robust against partial occlusions and discontinuity in video frames. It  outperforms the state-of-the-art methods in challenging datasets.  
$\tt{ROBUSfT}$ is implemented as a publicly available $\tt{C++}$ library and we provide a tutorial on how to use it in $\tt{https://github.com/mrshetab/ROBUSfT}$.
\end{abstract}


\IEEEpeerreviewmaketitle

\section{Introduction}
\begin{figure*}[h]
    \centering
    \includegraphics[height=18cm]{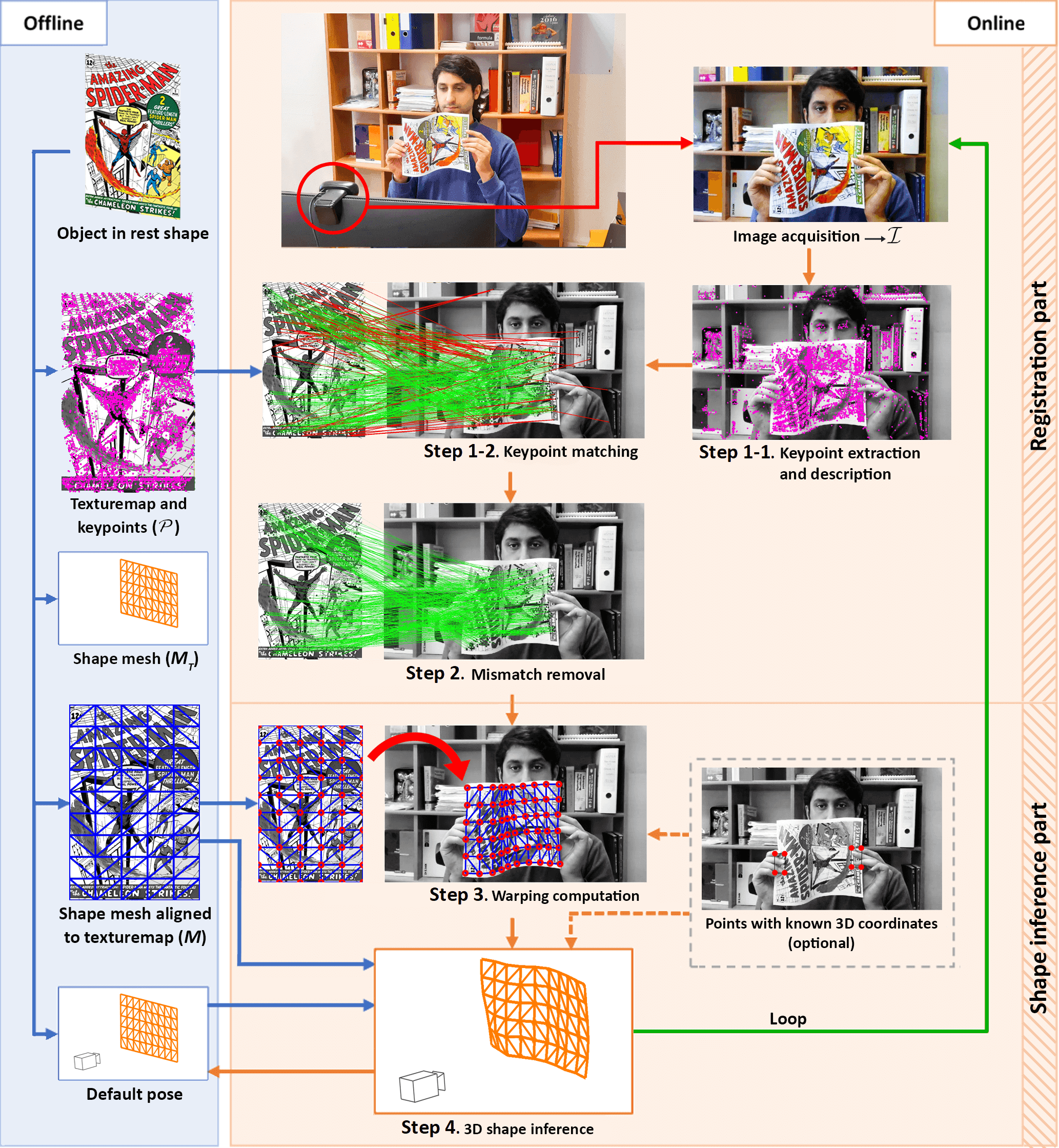}
    \caption{Overview  of  $\tt{ROBUSfT}$.}
    \label{pipeline_flowchart}
\end{figure*}

\noindent\textit{Problem and challenges.} Tracking the 3D shape of a deforming object has important applications in augmented reality~\cite{pilet2008fast,haouchine2014single}, computer-assisted surgery~\cite{hu2009non,collins2011deformable,malti2012template,collins2016robust,lamarca2020defslam} and robotics~\cite{li2014real,frank2014learning,arandamonocular}.
However, the existing solutions are impractical. This is because of the following challenges: (\texttt{C1}) real-time implementability and (\texttt{C2}) robustness. Challenge \texttt{C1} is hard to achieve because the solution usually involves a computationally demanding multi-step pipeline. Challenge \texttt{C2} is hard to maintain because of noises,  occlusions, invisible object, large deformations and fast motions. 
Furthermore, in numerous applications of augmented reality, computer-assisted surgery and robotics, a 2D camera is the de facto sensor owing to its light weight, small size, and low cost. The camera's perspective projection introduces an additional challenge, (\texttt{C3}) recoverability of shape's depth from a 2D image. Challenge \texttt{C3} becomes extremely difficult for deforming objects.

\vspace{.2cm}
\noindent\textit{Shape-from-Template.} 
Different priors and constraints have been proposed to resolve challenge \texttt{C3}. The most common ones are the object's 3D rest shape, texturemap, deformation law and the camera intrinsics. These form the ingredients for a variety of methods. Among these methods, we are particularly interested in Shape-from-Template (SfT). SfT has been well studied for isometrically  deforming objects~\cite{salzmann2009reconstructing,salzmann2008closed,perriollat2011monocular} and has been shown to uniquely resolve the depth of each object point~\cite{bartoli2015shape}. It uses a template formed by the abovementioned priors. SfT's input is a single image of a deformed object, and its output is the object's 3D shape seen in the image.  We adopt a conventional SfT pipeline shown in Figure~\ref{pipeline_flowchart} to solve the 3D shape tracking problem of deforming objects. The pipeline involves keypoint extraction and matching, mismatch filtering, warping and 3D shape inference steps, respectively. We successfully made it real-time and robust by integrating seamlessly both novel and state-of-the-art algorithms at different steps. We next overview the strengths and weaknesses of current SfT methods.

\vspace{.2cm}
\noindent\textit{State-of-the-art SfT methods.} SfT can be broken down into two main parts: registration and 3D shape inference. Following this, we categorize existing SfT methods into two groups:  (\texttt{G1}) shape inference methods and  (\texttt{G2}) integrated methods. \texttt{G1} methods only cover the 3D shape inference part~\cite{salzmann2008closed,salzmann2009reconstructing,bartoli2015shape,chhatkuli2016stable,famouri2018fast,brunet2014monocular,ozgur2017particle,arandamonocular}. In contrast, \texttt{G2} methods cover both the registration and 3D shape inference parts~\cite{salzmann2010linear,ostlund2012laplacian,ngo2015template,collins2015poster,collins2016robust,liu2017better}.
We also overview Deep Neural Network (DNN) based SfT methods, as the third group (\texttt{G3}), which have been recently introduced. \texttt{G3} methods cover both the registration and 3D shape inference parts~\cite{pumarola2018geometry,golyanik2018hdm,shimada2019ismo,FUENTESJIMENEZ2022104531,fuentes2021texture}.  
The majority of \texttt{G1} methods are wide-baseline. 
However, they barely run in real-time.
Furthermore, a complete solution with registration shall be even slower. 
The majority of \texttt{G2} methods require an initialization close to the solution. 
This makes them short-baseline. Subsequently they often fail against occlusions, fast motions and large deformations. Once failed, they need to be reinitialized. 
\texttt{G3} methods are wide-baseline and run in real-time. However, they are object-specific.
They require a huge amount of training data and proper computational resources for each new object. These make them difficult to consider as a general and ready-to-use solution.
We therefore conclude that there does not exist an SfT method that is complete, real-time, robust and easily applicable to new objects.

\vspace{.2cm}
\noindent\textit{Contributions.}
We list our contributions in three parts.
\paragraph{Contribution to SfT} 
We propose $\tt{ROBUSfT}$, a complete real-time robust SfT pipeline for monocular 3D shape tracking of isometrically deforming thin-shell objects with matchable appearance. 
It can track up to 30\,fps using $640\times480$ images on off-the-shelf standard consumer hardware.  It does not require initialization and implements tracking-by-detection. 
It is wide-baseline and robust to occlusions, invisible object, large deformations and fast motions. It does not require training. It is thus directly applicable in many industrial and research contexts. $\tt{ROBUSfT}$ outperforms the state-of-the-art methods in challenging datasets.

\paragraph{Contribution to mismatch removal} 
We introduce $\tt{myNeighbor}$, a novel mismatch removal algorithm. It handles deforming scenes and a large percentage of mismatches. It is lightning fast, reaching $200$\,fps.  

\paragraph{Contribution to experimental validation}
We design a novel type of validation procedure, called Fake Realistic Experiment ($\tt{FREX}$). It allows us to automatically generate semi-synthetic datasets with ground-truth. 
This eases the quantitative evaluation of 2D and 3D shape tracking algorithms for deforming objects to a great extent.  

\vspace{.2cm}
\noindent\textit{Paper structure.} Section 2 reviews previous work. Section 3 explains $\tt{ROBUSfT}$. Section 4 presents $\tt{FREX}$. Section 5 describes $\tt{myNeighbor}$, conducts a series of experiments and evaluates the results of $\tt{myNeighbor}$ in comparison to previous work. 
Section 6 validates $\tt{ROBUSfT}$ through $\tt{FREX}$ and real data experiments, and compares the results with previous work. 
Finally, Section 7 concludes and suggests future work.

\section{Previous Work}
We review the methods for monocular shape inference of isometrically deforming objects, following the above three categories, namely, 
(\texttt{G1}) shape inference methods, (\texttt{G2}) integrated methods, and (\texttt{G3}) DNN-based SfT methods.
For each category, we describe the assumptions, main characteristics, and limitations.
We finally compare $\tt{ROBUSfT}$ to these methods. 

\subsection{(\texttt{G1}) Shape inference methods}
\label{subsubsec:Shapeinferencemethods}
These methods cover the 3D shape inference part. 
They assume that the registration between the template and the image was previously computed. 
For instance, they typically use keypoint matches between the template and the image, with generic mismatch removal methods~\cite{pizarro2012feature,collins2014analysis,tran2012defence,pizarro2012feature,famouri2018fast}.
In fact, very few methods in this category could form a complete SfT pipeline by adding an existing registration solution~\cite{pilet2008fast,famouri2018fast}.
Three general groups are found in existing 3D shape inference methods; \textit{(i)} methods using a convex relaxation of isometry called inextensibility~\cite{salzmann2008closed,salzmann2009reconstructing,brunet2014monocular}, \textit{(ii)} methods using local differential geometry~\cite{bartoli2015shape,chhatkuli2016stable,famouri2018fast}, and  \textit{(iii)} methods minimizing a global non-convex cost function~\cite{brunet2014monocular,ozgur2017particle,arandamonocular}. 
The methods in \textit{(iii)} are the most precise ones but also computationally expensive and require initialization. 
The first two groups of methods are often used to provide an initial guess for the third group.

In the first group, Salzmann et al.~\cite{salzmann2008closed} suggested a closed-form solution to non-rigid 3D surface registration by solving a set of quadratic equations accounting for inextensibility. Later, they replaced equality constraints with inequality and thus sharp deformations could be better recovered~\cite{salzmann2009reconstructing}. Brunet et al.~\cite{brunet2014monocular} formulated two shape inference methods based on point-wise and continuous surface models as Second Order Cone Programs (SOCP).
In the second group, Bartoli et al.~\cite{bartoli2015shape} showed that in addition to keypoint 2D coordinates in the image, their first-order differential structure can be used to estimate the depth. 
Instead of calculating the warp globally, which is time-consuming, Famouri et al.~\cite{famouri2018fast} estimated the depth locally for each match pair with respect to both local texture and neighboring matches. 
In each frame, the most recognizable matches were selected based on offline training. The execution speed of their algorithm is claimed to be up to 14\,fps only for the 3D shape inference. 
In the third group, Brunet et al.~\cite{brunet2014monocular} proposed a refining isometric SfT method by reformulating the isometric constraint and solving as a non-convex optimization problem.
The method required a reasonably accurate 3D shape of the deforming surface as the initializing guess.
\"Ozg\"ur and Bartoli~\cite{ozgur2017particle}, developed Particle-SfT, which  handles isometric and non-isometric deformations. A particle system is guided by deformation and reprojection constraints which are applied consecutively to the particle mesh. Similar to~\cite{brunet2014monocular}, this algorithm needs an initial guess for the 3D position of the particles, however, for~\cite{ozgur2017particle}, sensitivity to this initial guess is very low. The closer the guess to the true 3D shape, the faster the convergence.
Aranda et al.~\cite{arandamonocular} improved this algorithm in terms of execution speed and occlusion-resistance and used that in real-time shape servoing of isometrically deforming objects. They used the 3D shape estimated in one frame as the initial guess for the next frame and thus improved the convergence speed of the algorithm to a great extent. They showed that their algorithm can track a paper sheet covered with markers and being manipulated by a robotic arm. To this end, they only needed to track a handful of markers. Knowing the 3D coordinates of several mesh points also has a significant effect on the convergence speed of the algorithm. The last step of $\tt{ROBUSfT}$ uses the same method to infer the 3D shape, as explained in Section III.


\subsection{(\texttt{G2}) Integrated methods}
\label{subsubsec:Integratedmethods}
These methods handle registration and 3D shape inference at the same time. They minimize a non-convex cost function in order to align the 3D inferred shape with image features. These features can be local~\cite{ostlund2012laplacian,ngo2015template} or at the pixel-level~\cite{collins2015poster,collins2016robust}. 

Ostlund et al.~\cite{ostlund2012laplacian} and later Ngo et al.~\cite{ngo2015template} used the Laplacian formulation to reduce the problem size by introducing control points on the surface of the deforming object. The process of removing mismatches was performed iteratively during optimization by projecting the 3D estimated shape on the image and disregarding the correspondences with higher reprojection errors.
Using this procedure, they could reach up to 10\,fps using $640\times480$ input images and restricting the maximum number of template and image keypoints to 500 and 2000, respectively.

As for pixel-level alignment, Collins and Bartoli~\cite{collins2015poster} introduced a real-time SfT algorithm which could handle large deformations and occlusions and reaches up to 21\,fps. They combined extracted matches with physical deformation priors to perform shape inference. Collins et al.~\cite{collins2016robust} later extended this algorithm and used it for tracking organs in laparoscopic videos. For achieving better performance, they also exploited organ boundaries as a tracking constraint. 

These methods are fast and can handle large deformation. Their main drawback, however, is to be short-baseline. In case of tracking failure, they should be re-initialized precisely with a wide-baseline method.
This restrict their usage to video streams.



\begin{table*}[h]
\centering
    \begin{tabular}{ccccccccc}
        \toprule
	    Category & Method & Registration & Real-time & Wide-baseline & \makecell{General \\ geometry} & \makecell{Needless of \\ training for \\ new objects} & \makecell{Public access \\ code} \\
      	\hline
        \multicolumn{1}{c|}{\multirow{6}{*}{{\tt G1}}} & 
        \multicolumn{1}{c}{Salzmann et al.~\cite{salzmann2008closed}} & 
        \multicolumn{1}{c}{$\times$} & 
        \multicolumn{1}{c}{NA} & 
        \multicolumn{1}{c}{$\checkmark$} & 
        \multicolumn{1}{c}{$\checkmark$} & 
        \multicolumn{1}{c}{$\checkmark$} & 
        \multicolumn{1}{c}{$\times$} \\
        \multicolumn{1}{c|}{} & 
        \multicolumn{1}{c}{Brunet et al.~\cite{brunet2014monocular}} & 
        \multicolumn{1}{c}{$\times$} & 
        \multicolumn{1}{c}{$\times$} & 
        \multicolumn{1}{c}{$\checkmark$} & 
        \multicolumn{1}{c}{$\checkmark$} & 
        \multicolumn{1}{c}{$\checkmark$} & 
        \multicolumn{1}{c}{$\checkmark$} \\
        \multicolumn{1}{c|}{} & 
        \multicolumn{1}{c}{Bartoli et al.~\cite{bartoli2015shape}} & 
        \multicolumn{1}{c}{$\times$} &
        \multicolumn{1}{c}{NA} & 
        \multicolumn{1}{c}{$\checkmark$} & 
        \multicolumn{1}{c}{$\checkmark$} & 
        \multicolumn{1}{c}{$\checkmark$}  & 
        \multicolumn{1}{c}{$\checkmark$} \\
	    \multicolumn{1}{c|}{} & 
	    \multicolumn{1}{c}{Ozgur et Bartoli~\cite{ozgur2017particle}}  & 
	    \multicolumn{1}{c}{$\times$} & 
	    \multicolumn{1}{c}{$\times$} & 
	    \multicolumn{1}{c}{$\checkmark$} & 
	    \multicolumn{1}{c}{$\checkmark$} & 
	    \multicolumn{1}{c}{$\checkmark$} & 
	    \multicolumn{1}{c}{$\times$}  \\
	    \multicolumn{1}{c|}{} & 
	    \multicolumn{1}{c}{Famouri et al.~\cite{famouri2018fast}} & 
	    \multicolumn{1}{c}{$\times$} & 
	    \multicolumn{1}{c}{$\checkmark$} & 
	    \multicolumn{1}{c}{$\checkmark$} & 
	    \multicolumn{1}{c}{$\checkmark$} & 
	    \multicolumn{1}{c}{$\checkmark$} & 
	    \multicolumn{1}{c}{$\checkmark$} \\
	    \multicolumn{1}{c|}{} & 
	    \multicolumn{1}{c}{Aranda et al.~\cite{arandamonocular}} & 
	    \multicolumn{1}{c}{$\times$} & 
	    \multicolumn{1}{c}{$\checkmark$} & 
	    \multicolumn{1}{c}{$\checkmark$} & 
	    \multicolumn{1}{c}{$\checkmark$} & 
	    \multicolumn{1}{c}{$\checkmark$} & 
	    \multicolumn{1}{c}{$\times$}  \\\hline

	    \multicolumn{1}{c|}{\multirow{4}{*}{{\tt G2}}} & 
	    \multicolumn{1}{c}{Ostlund et al.~\cite{ostlund2012laplacian}} & 
	    \multicolumn{1}{c}{$\checkmark$} & 
	    \multicolumn{1}{c}{$\checkmark$} & 
	    \multicolumn{1}{c}{$\times$} & 
	    \multicolumn{1}{c}{$\checkmark$} & 
	    \multicolumn{1}{c}{$\checkmark$} & 
	    \multicolumn{1}{c}{$\times$} \\
        \multicolumn{1}{c|}{} & 
        \multicolumn{1}{c}{Ngo et al.~\cite{ngo2015template}} & 
        \multicolumn{1}{c}{$\checkmark$} & 
        \multicolumn{1}{c}{$\checkmark$} & 
        \multicolumn{1}{c}{$\times$} & 
        \multicolumn{1}{c}{$\checkmark$} & 
        \multicolumn{1}{c}{$\checkmark$} & 
        \multicolumn{1}{c}{$\times$} \\
        \multicolumn{1}{c|}{} & 
        \multicolumn{1}{c}{Collins and Bartoli~\cite{collins2015poster}} & 
        \multicolumn{1}{c}{$\checkmark$} & 
        \multicolumn{1}{c}{$\checkmark$} & 
        \multicolumn{1}{c}{$\times$} & 
        \multicolumn{1}{c}{$\checkmark$} & 
        \multicolumn{1}{c}{$\checkmark$} & 
        \multicolumn{1}{c}{$\times$} \\
        \multicolumn{1}{c|}{} & \multicolumn{1}{c}{Collins et al.~\cite{collins2016robust}} &
        \multicolumn{1}{c}{$\checkmark$} & 
        \multicolumn{1}{c}{$\checkmark$} & 
        \multicolumn{1}{c}{$\times$} & 
        \multicolumn{1}{c}{$\checkmark$} & 
        \multicolumn{1}{c}{$\checkmark$} & 
        \multicolumn{1}{c}{$\times$} \\\hline
        
	    \multicolumn{1}{c|}{\multirow{5}{*}{{\tt G3}}} & 
	    \multicolumn{1}{c}{Pumarola et al.~\cite{pumarola2018geometry}} & 
	    \multicolumn{1}{c}{$\checkmark$} & 
	    \multicolumn{1}{c}{$\times$} & 
	    \multicolumn{1}{c}{$\checkmark$} & 
	    \multicolumn{1}{c}{$\times$} & 
	    \multicolumn{1}{c}{$\times$} & 
	    \multicolumn{1}{c}{$\times$} \\
        \multicolumn{1}{c|}{} & 
        \multicolumn{1}{c}{Golyanik et al.~\cite{golyanik2018hdm}} & 
        \multicolumn{1}{c}{$\checkmark$} & 
        \multicolumn{1}{c}{$\checkmark$} & 
        \multicolumn{1}{c}{$\checkmark$} & 
        \multicolumn{1}{c}{$\times$} & 
        \multicolumn{1}{c}{$\times$} & 
        \multicolumn{1}{c}{$\times$} \\
        \multicolumn{1}{c|}{} &
        \multicolumn{1}{c}{Fuentes-Jimenez et al.~\cite{FUENTESJIMENEZ2022104531}} & 
        \multicolumn{1}{c}{$\checkmark$} & 
        \multicolumn{1}{c}{$\checkmark$} & 
        \multicolumn{1}{c}{$\checkmark$} & 
        \multicolumn{1}{c}{$\checkmark$} & 
        \multicolumn{1}{c}{$\times$} & 
        \multicolumn{1}{c}{$\times$} \\
        \multicolumn{1}{c|}{} & 
        \multicolumn{1}{c}{Shimada et al.~\cite{shimada2019ismo}} & 
        \multicolumn{1}{c}{$\checkmark$} & 
        \multicolumn{1}{c}{$\checkmark$} & 
        \multicolumn{1}{c}{$\checkmark$} & 
        \multicolumn{1}{c}{$\times$} & 
        \multicolumn{1}{c}{$\times$} & 
        \multicolumn{1}{c}{$\times$} \\
        \multicolumn{1}{c|}{} &
        \multicolumn{1}{c}{Fuentes-Jimenez et al.~\cite{fuentes2021texture}} & 
        \multicolumn{1}{c}{$\checkmark$} & 
        \multicolumn{1}{c}{$\checkmark$} & 
        \multicolumn{1}{c}{$\checkmark$} & 
        \multicolumn{1}{c}{$\times$} & 
        \multicolumn{1}{c}{$\checkmark$} & 
        \multicolumn{1}{c}{$\times$} \\\hline
        
        \multicolumn{1}{c}{} & 
        \multicolumn{1}{c}{$\tt{ROBUSfT}$} & 
        \multicolumn{1}{c}{$\checkmark$} & 
        \multicolumn{1}{c}{$\checkmark$} & 
        \multicolumn{1}{c}{$\checkmark$} & 
        \multicolumn{1}{c}{$\checkmark$} & 
        \multicolumn{1}{c}{$\checkmark$} & 
        \multicolumn{1}{c}{$\checkmark$} \\ \hline
        \end{tabular}
    \caption{Comparison of the state-of-the-art SfT methods and $\tt{ROBUSfT}$.}
    \label{tab:methods_comparison}
\end{table*}

\subsection{(\texttt{G3}) DNN-based methods}
\label{subsubsec:DNNbasedmethods}
DNN-based SfT methods have been introduced in the recent years, which coincides with the tendency to use deep learning to solve many computer vision problems. 
These methods are wide-baseline, fast, and cover both the registration and shape inference steps~\cite{pumarola2018geometry,golyanik2018hdm,shimada2019ismo,FUENTESJIMENEZ2022104531,fuentes2021texture}. 
We group these methods based on their type of output, which may be sparse or dense.  
The methods of the first group represent the SfT solution as the 3D coordinates of a regular mesh with a predefined size~\cite{pumarola2018geometry,golyanik2018hdm,shimada2019ismo}.
The usage of these methods is limited to thin-shell objects with rectangular shapes. 
The second group of methods gives a pixel-level depthmap as output~\cite{FUENTESJIMENEZ2022104531,fuentes2021texture}. 
They also apply a post-processing step based on the As-rigid-as-possible (ARAP) model~\cite{sorkine2007rigid} to the resulting depthmap. This step recovers the whole  object, including the occluded parts, as a mesh.
The method in~\cite{FUENTESJIMENEZ2022104531} reconstructs the shape of the object with different geometries and texturemaps that the network is trained for. In~\cite{fuentes2021texture}, however, the proposed method can be applied to objects with new texturemaps unseen to the network. The geometry of the objects is, nevertheless, limited to flat paper-like shapes.
All the aforementioned methods in this category are object-specific. This means that they merely work for the object that they were trained for. An exception is~\cite{fuentes2021texture}, as it works for unseen texturemaps but the applicability is still limited to flat rectangular objects. On the other hand, in order to use the DNN-based methods for a new object, the network should be fine-tuned for it. This demands proper computational resources and potentially a huge amount of training data, which are challenging to collect for deformable objects.  

\subsection{Positioning $\tt{ROBUSfT}$ compared to previous work}
Existing methods all have one or several limitations, including not covering the whole pipeline, not being wide-baseline, being limited to specific texture or geometry, requiring fine-tuning for a new object, being slow, and lacking public code access. This information is summarized in Table~\ref{tab:methods_comparison}.
In contrast, $\tt{ROBUSfT}$ covers the whole pipeline and due to the fast execution can be used to develop real-time shape tracking applications. It can be instantly used for each deforming object without training. Only a template containing information regarding the object’s geometry, appearance, and deformation law as well as intrinsic parameters of the monocular camera is necessary, but this need is common to all existing and future SfT methods, by definition. In the next section, we describe $\tt{ROBUSfT}$ and all its steps.

\section{$\tt{ROBUSfT}$}   \label{sec:ROBUSfT}
\subsection{Overview of the pipeline} 

The overview of our pipeline is presented in Figure~\ref{pipeline_flowchart}.
The pipeline is divided into an offline and an online sections. The offline section deals with the template. The online section includes four main steps: keypoint extraction and matching, mismatch removal, warp estimation, and 3D shape inference. The images coming directly from the camera are used as the inputs for the first step. In this step, the keypoints are extracted and matched with the ones that were previously extracted from the template's texturemap. Then, the mismatches are detected and removed using our new mismatch removal algorithm $\tt{myNeighbor}$. The list of estimated correct matches is then transferred to the next step where a warp is estimated between the template's texturemap and the image. This warp transfers the template's registered mesh to the image space, which is finally used as input for the 3D shape inference algorithm. This process is repeated for each image, the analysis of each image being performed independently in a tracking-by-detection manner.

In the following, both the offline and online sections of the pipeline are described in detail. Afterwards, an implementation permitting a fast execution of the pipeline is given.

\subsection{Offline section: creating a template} 
We create a template for the surface of the deforming object that we want to track. We call this surface the tracking surface. The template of the tracking surface consists of the following elements: 
\begin{itemize}
    \item ${M_T}$: the triangular mesh covering the tracking surface at rest shape.
    \item $\mathcal{P}$: the texturemap of the tracking surface.
    \item ${M}$: the alignment of ${M_T}$ to $\mathcal{P}$.
\end{itemize}
The first step in creating the template is to generate the 3D model of the tracking surface. The 3D model is in fact the textured 3D geometry of the tracking surface in real dimensions in rest shape. 
We form ${M_T}$ by triangulating this 3D geometry. The resolution of ${M_T}$ should be high enough to be well aligned to the shape of the tracking surface.
The next step is to take an image from the 3D model of the tracking surface while it is positioned perpendicular to the camera's optical axis in a simple texture-less background. In this image, $\mathcal{P}$ is formed by the projection of the texture of the tracking surface and ${M}$ by the projection of ${M_T}$.
For simple rectangular thin-shell objects like a piece of paper, the whole process is straightforward. For other objects, including thin-shell objects with arbitrary shape, such as a shoe sole, and also volumetric objects, 3D reconstruction software like Agisoft Photoscan~\cite{Agisoft} can be used.

Next, we extract keypoints on $\mathcal{P}$. These keypoints will be matched with the ones that will be extracted from the input image in the online section. We use SIFT~\cite{lowe2004distinctive} for extracting keypoints but any other feature descriptor could be swapped in. 
As the final step, we initialize the pose of ${M_T}$ in 3D space. This initial pose can be arbitrarily chosen as it will be used only once by Step 4 of the online section of the pipeline for the first input image. It will then be replaced by the inferred 3D shape in the next images. 

In order to use the $\tt{ROBUSfT}$ C++ library, first, an object of the class $\tt{ROBUSfT}$ should be created. The whole process of forming the template for this object is handled by the member function $\tt{build\_template()}$. This function possesses parameters for creating templates for rectangular and non-rectangular thin-shell objects as well as the tracking surface of volumetric objects. Regarding thin-shell objects, the process of forming the template is automatic by just receiving a handful of inputs from the user. For the tracking surface of volumetric objects, however, ${M_T}$, ${M}$, and $\mathcal{P}$ should be prepared by the user and imported into the library. 

\subsection{Online section: shape tracking} 

\textit{Step 1: keypoint extraction and matching.} The first step of the online section of the pipeline is to extract keypoints in the input image $\mathcal{I}$. To do so, we use the PopSift library~\cite{Griwodz2018Popsift}, which is a GPU implementation of the SIFT algorithm. We then match these keypoints with the ones that were previously extracted from $\mathcal{P}$ by comparing descriptors, using winner-takes-all and Lowe's ratio test.
Inevitably, a number of mismatches will be formed between $\mathcal{P}$ and $\mathcal{I}$. The mismatch points in $\mathcal{I}$ can be located on the surface of the deforming object or even in the background. This is shown as red lines in the \textit{Matching} step of Figure~\ref{pipeline_flowchart}. These mismatches will be eliminated in \textit{Step 2} thanks to $\tt{myNeighbor}$ which can cope with a large percentage of mismatches. As a result, in this step, the images coming from the camera can be used directly without pretraining on either the image for segmenting the object from the background, or the matches for preselection of the most reliable ones. 
In the library, the member function $\tt{extract\_keypoints\_GPU()}$ handles the keypoint extraction in $\mathcal{I}$. Then, the member function $\tt{match()}$ performs matching. 

\vspace{5pt}
\textit{Step 2: mismatch removal.} To remove the possible mismatches introduced in \textit{Step 1}, a new mismatch removal algorithm, $\tt{myNeighbor}$, was developed. The main principle used in this algorithm is the preservation of the neighborhood structure of correct matches on a deforming object. In other words, if all of the matches were correct, by deforming the object, the neighbor matches of each match should be preserved. On the contrary, mismatches lead to differences in the neighboring matches of each matched point in $\mathcal{I}$ in comparison to $\mathcal{P}$. This was used as a key indication to detect and remove mismatches. The whole process of $\tt{myNeighbor}$ is explained in Section V.
In the library, the member function $\tt{mismatch\_removal\_algorithm()}$ handles the mismatch removal process. The output is a list of estimated correct matches.

\begin{figure}
    \centering
    \includegraphics[width=\linewidth]{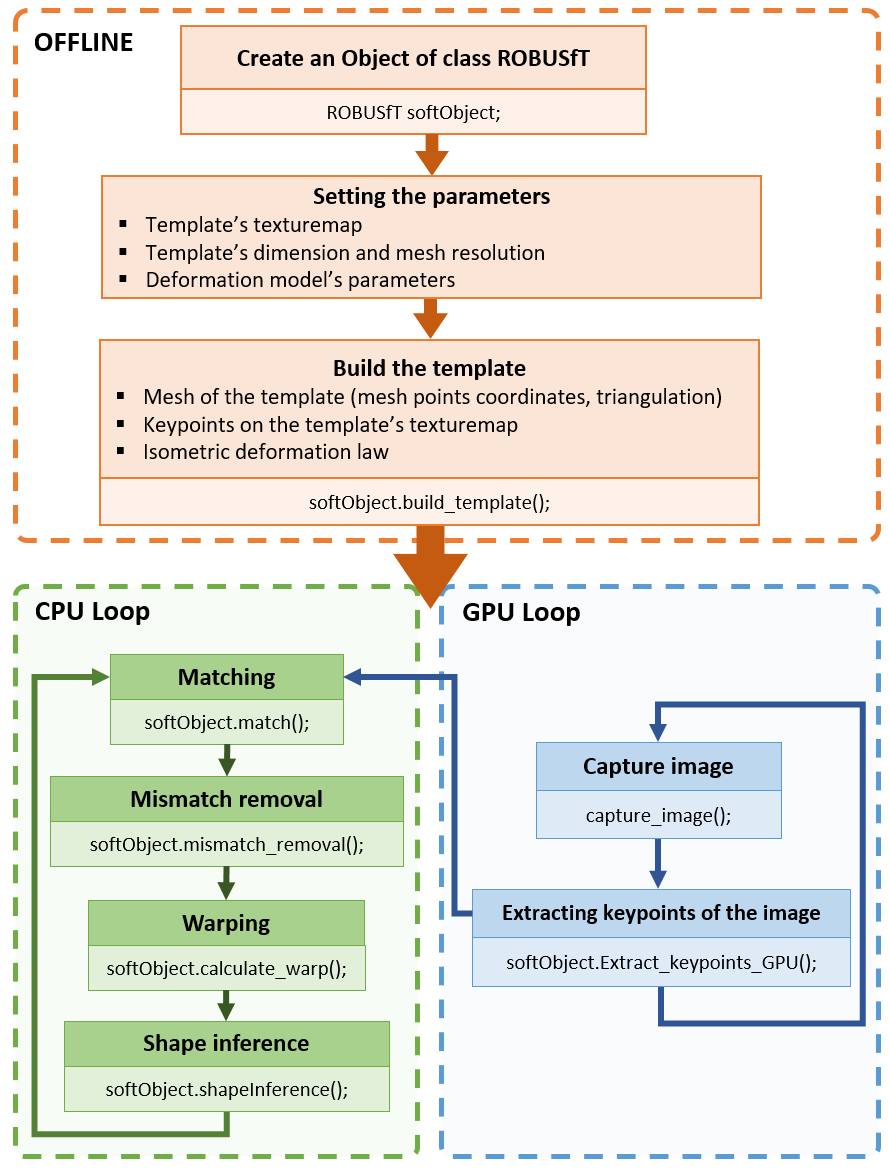}
    \caption{Implementation of $\tt{ROBUSfT}$ on the CPU and GPU. A pure CPU implementation is also available.}
    \label{image_implementation}
\end{figure}
\vspace{5pt}
\textit{Step 3: warp estimation.} We use the estimated correct matches to estimate a warp $W$ between $\mathcal{P}$ and $\mathcal{I}$. We then use $W$ to transfer ${M}$ to $\mathcal{I}$ and form ${\widehat{M}}$. The mesh points in ${\widehat{M}}$ will be used as sightline constraints in the 3D shape inference algorithm in \textit{Step 4}. 
\begin{figure*}
    \centering
    \includegraphics[width=\linewidth]{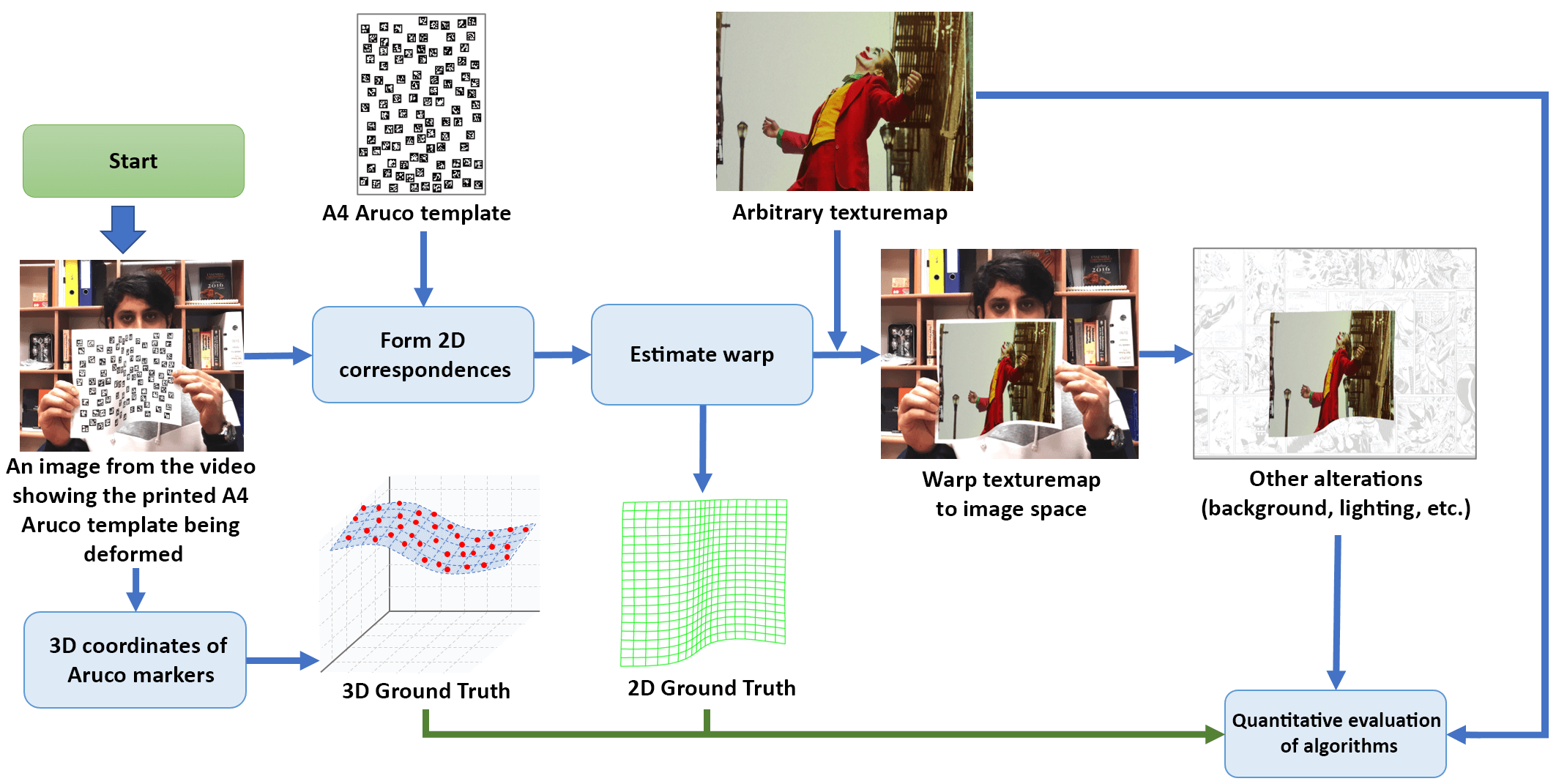}
    \caption{Flowchart of $\tt{FREX}$.}
    \label{image_fake_realistic_experiment_flowchart}
\end{figure*}
The precision of warping depends on the number of matches, their correctness, and their distribution all over $\mathcal{P}$. Warp $W$ can be estimated in the most precise way if all the matches are correct between $\mathcal{P}$ and $\mathcal{I}$. However, due to the smoothing nature of the warping algorithms, the transferring process can cope with a small percentage of mistakenly selected mismatches.
It should be noted that $W$ cannot be extremely precise in areas without matches. As a result, in these areas, the shape of ${\widehat{M}}$ might not be aligned well to the shape of the deforming object in $\mathcal{I}$. This is worse when the matchless area is located near the boundaries of $\mathcal{P}$ as the alignment cannot be guided by the surrounding matches. 
Hence, in order to use just well-aligned transferred mesh points of ${\widehat{M}}$ as the input for the 3D shape inference step, an assessment is performed over all of the mesh points and only the qualified ones are passed to \textit{Step 4}. 
For this, we check ${M}$ cell-by-cell. Only the mesh vertices for cells containing at least one correct match will be qualified as salient mesh points. The indices of these mesh points and their coordinates in ${\widehat{M}}$ are passed to \textit{Step 4}. The other mesh points are disregarded.

Representing and estimating $W$ can be done with two well-known types of warp, the Thin-Plate Spline (TPS)~\cite{bookstein1989principal} and the Bicubic B-Spline (BBS) warps~\cite{rueckert1999nonrigid}, which we both tested. The former is based on radial basis functions while the latter is formulated on the tensor-product.
Having the same number of matches as input, the TPS warp proved to be more precise than the BBS warp; nevertheless, its execution time rises exponentially with increasing number of matches. The execution time, however, remains almost constant for the BBS warp regardless of the number of matches. Thus, considering the criterion of fast execution of the code, the BBS warp was chosen as the warp function in this step and also in the mismatch removal step discussed in Section V.
In the library, the process of warp estimation is performed by the function $\tt{warp()}$ that calls two functions $\tt{BBS\_Function()}$ and $\tt{BBS\_Evaluation()}$. The former estimates the warp $W$ while the latter uses $W$ to transfer ${M}$ and form ${\widehat{M}}$. The process of selecting the salient mesh points is done by the member function $\tt{set\_sightlines()}$.

\vspace{5pt}
\textit{Step 4: 3D shape inference.} 
We use Particle-SfT~\cite{ozgur2017particle} as improved for tracking in~\cite{arandamonocular}. 
In this algorithm, a particle system is defined from the points and edges in ${M_T}$. Then, the sightline and deformation constraints are applied consecutively on the particles until they converge to a stable 3D shape. As described in~\cite{arandamonocular}, in order to increase the convergence speed of the algorithm, the stable 3D shape for an image is used as initial guess for the next image. It should be noted that Particle-SfT can work even without a close initial guess. If the object is invisible in one or several images, the last inferred 3D shape can be used as the initial guess for the upcoming frame containing the object. 
This results in a slightly longer computation time in that image. For the next upcoming images the normal computation time is resumed. 
This capability brings about two of the major advantages of our pipeline, which are being wide-baseline and robust to video discontinuities. 
In the library, the whole process of shape inference is handled by the member function $\tt{shapeInference()}$.

As mentioned in~\cite{arandamonocular}, one of the optional input data that can significantly improve the convergence of Particle-SfT is the existence of 3D known coordinates of one or several particles. This is shown in Figure~\ref{pipeline_flowchart}. The known 3D coordinates can be fixed in space, or can move on a certain trajectory. The latter happens when the deforming object is manipulated by tools with known poses in 3D space like robotic end-effectors.

\subsection{Implementation}  \label{subsec:Implementation}

In order to optimize the implementation of $\tt{ROBUSfT}$, it was coded in C++ in two parallel loops: one on the GPU, and one on the CPU. The GPU loop handles keypoint extraction in the images. These keypoints are transferred to the CPU loop where the rest of the steps of the pipeline are taken. 
A pure CPU implementation is also available.
This is shown in Figure~\ref{image_implementation}.
Any arbitrary resolution can be considered for the captured images, nevertheless, we obtained the best performance by using $640\times480$ images. The code runs on a Dell laptop with an Intel Core i7 2.60\,GHz CPU and a Quadro T1000 GPU.


\section{Fake Realistic Experiment ($\tt{FREX}$)} \label{sec:Fake realistic experiment}
We introduce a novel experimental protocol, which we used for evaluating $\tt{myNeighbor}$ and $\tt{ROBUSfT}$ in comparison to the state-of-the-art methods.
A single execution of this protocol provides a large collection of scenes of an isometrically deforming object in various conditions, with known 2D and 3D ground truth.
This collection can be used to evaluate, compare, train, and validate new algorithms regarding isometrically deforming objects such as mismatch removal, 2D image registration, and isometric 3D shape inference.
In contrast to other artificially generated scenes of an isometrically deforming surface, the generated images in our protocol are the result of real object deformations. Being formed of successive images with continuous deformation, it can also be used for algorithms which exploit feature and shape tracking. In addition, object occlusion and invisibility can be easily simulated, by dropping frames or pasting an occluder.

The protocol flowchart is shown in Figure~\ref{image_fake_realistic_experiment_flowchart}. First, we form the \textit{Aruco template} by randomly distributing a set of Aruco markers all over a blank image. We then print the Aruco template on a standard A4 paper.
These markers should be big enough to be recognizable by the user's camera in the desired distance.
In order to improve recognition, there should be white space between the markers on the paper. In our experiments, we used 100 markers with a width of 1.4\,cm. The OpenCV library was used to identify the markers. These markers were recognizable by a 720p RGB camera from an approximate distance of 0.6m. 
The next step is to deform the printed Aruco template in front of the camera. In each frame, the 2D and 3D coordinates of the markers' centers are estimated.
Because each marker has its own unique id, they can be used as correspondences between the Aruco template and each image of the video. 
We exploit the 2D coordinates of these recognized correspondences to estimate a warp with which we can transfer an arbitrary texturemap to the video image space.
This is done firstly by resizing the arbitrary texture to the size of the Aruco template. In order to keep the aspect ratio of the arbitrary texturemap, white margins can also be added before resizing.
Then, an inverse warping process with bilinear interpolation is used to transfer the pixel color information from the arbitrary texturemap to their corresponding pixels in the video images. 
The whole procedure results in a scene with the arbitrary texturemap being deformed exactly on top of the Aruco template.
It is also possible to add further modifications; for instance, one
can transfer the arbitrary texturemap to another scene with any different background. Besides, as in~\cite{varol2012monocular}, an artificial lighting can also be added to form different variations of the scene. 

For evaluating algorithms, one can use the 2D and 3D ground truth estimated in each frame of the video.
Regarding the 2D ground truth, the estimated warp can be used to identify the 2D corresponding point of each pixel of the arbitrary texturemap in the image.   
As for the 3D ground truth, one can exploit the 3D estimated coordinates of the Aruco markers in each frame which can be achieved using the OpenCV library.

\section{$\tt{myNeighbor}$}
We describe $\tt{myNeighbor}$, our novel mismatch removal algorithm. It works based on two main principles:
\begin{itemize}
    \item Having an image of a textured surface and another image of that surface undergoing a deformation, to estimate  a sufficiently accurate transfer function between them with which one can judge the correctness of matches, there is no need to remove all the mismatches from the list of matches. Instead, a set of correct matches would be sufficient to estimate the transfer function.
    \item This set of correct matches can be extracted considering that in reality, under a deformation, the neighborhood structure among the points on a deforming surface is preserved. 
\end{itemize}

\begin{figure}
    \centering
    \includegraphics[width=\linewidth]{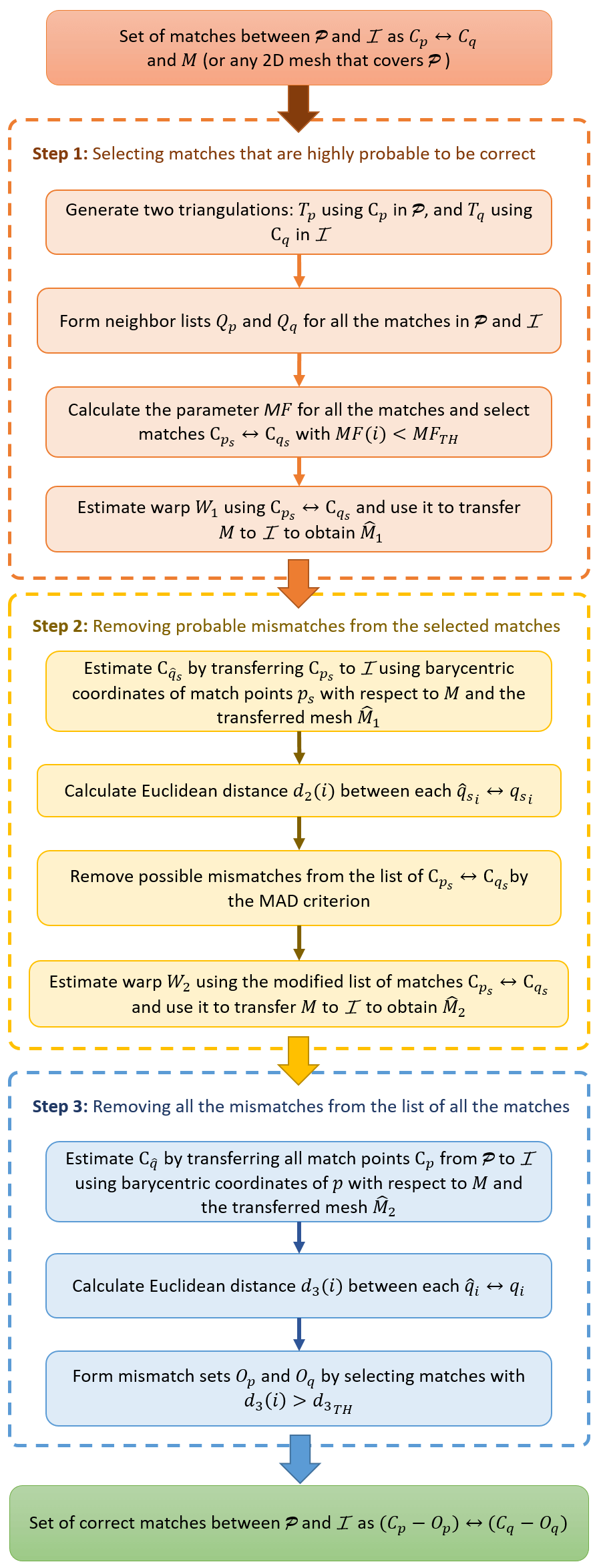}
    \caption{Flowchart of $\tt{myNeighbor}$.}
    \label{image_outlier_flowchart}
\end{figure}

\begin{figure*}
    \centering
    \includegraphics[width=\linewidth]{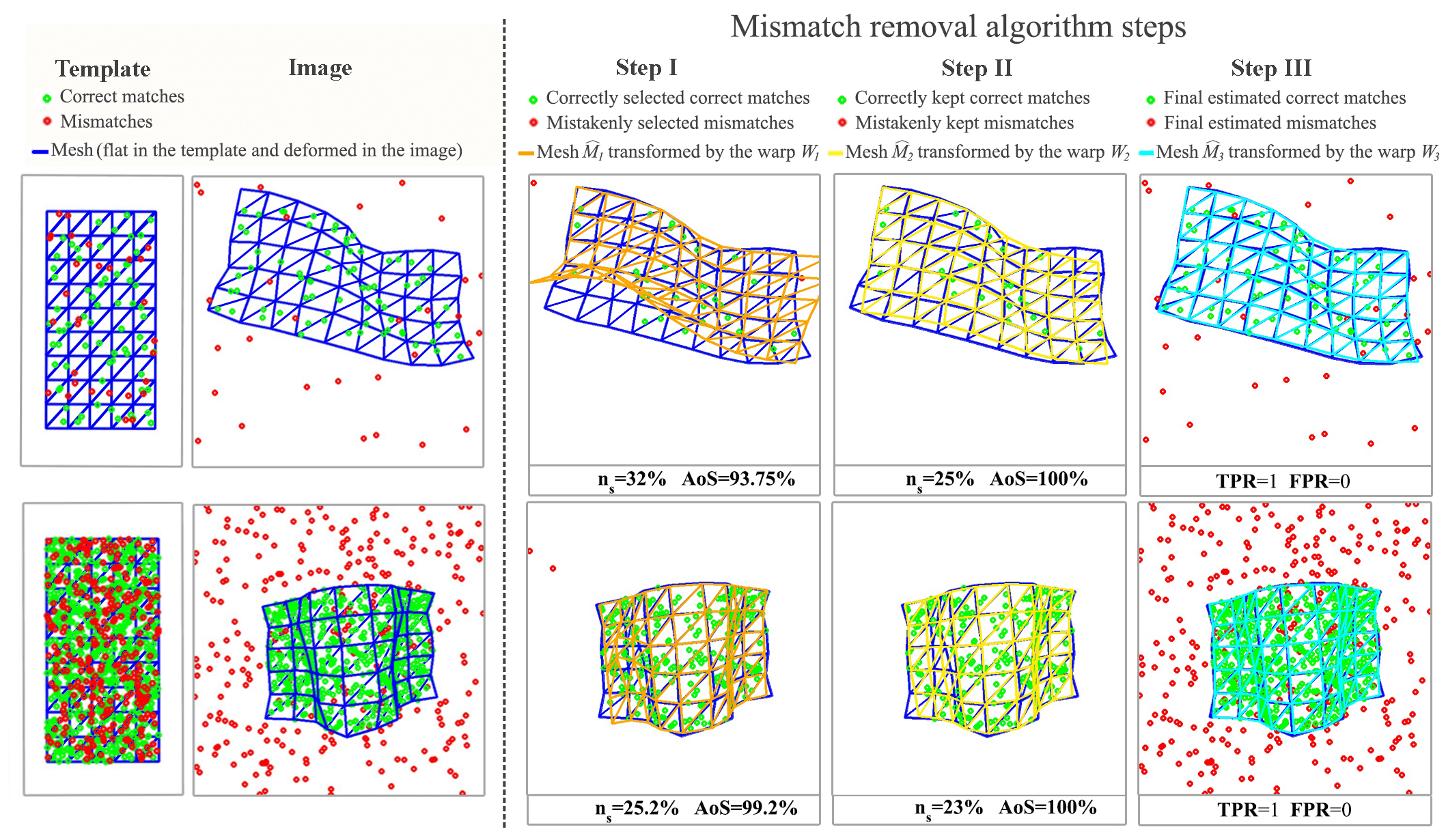}
    \caption{Two sample results of the steps for synthetic data experiments. The first row is an experiment with 100 matches and a mismatch percentage of 30\%. The second row is an experiment with 1000 matches and a mismatch percentage of 30\%. The first and second columns represent $\mathcal{I}_{F}$ and $\mathcal{I}$ with correct matches in green and mismatches in red. The third column is the result of \textit{Step I}. The wrongly chosen mismatches are shown in red. The fourth column is the result of \textit{Step II}. The mismatches along with a small percentage of correct matches are removed. The fifth column is the separation of the estimated correct matches and the estimated mismatches from \textit{Step III}. The transferred meshes ${\widehat{M}_1}$, ${\widehat{M}_2}$, and ${\widehat{M}_3}$ are shown in orange, yellow, and cyan for the three steps.}
    \label{image_outlier_removal_syn}
\end{figure*}

We show that by using these two principles, the mismatches can be detected and removed in a fast and efficient way. 
The proposed algorithm is illustrated in Figure~\ref{image_outlier_flowchart}. 
It consists of three steps.
First, a set of matches which are highly probable to be correct are selected. This selection is done by forming two triangulations using match points, one in $\mathcal{P}$ and one in $\mathcal{I}$, and then choosing matches with high similarity in the list of their neighbors. Second, a small percentage of possible mismatches among the selected matches are identified and removed.
This is done by transferring the selected match points from $\mathcal{P}$ to $\mathcal{I}$ and then removing those with large distances from their correspondences in $\mathcal{I}$. Third, we transfer all the match points from $\mathcal{P}$ to $\mathcal{I}$ using a warp estimated based on the clean set of selected matches from the second step. The distance between the transferred template match points and their correspondences in $\mathcal{I}$ is used as the criterion to distinguish estimated mismatches from estimated correct matches\color{black}. 

In order to analyze the performance of $\tt{myNeighbor}$ and calibrate the parameters in the different steps, we used synthetic data experiments. In the following section, we describe the design of these experiments. Afterwards, we describe in detail the different steps of $\tt{myNeighbor}$. 

\begin{figure}
    \centering
    \includegraphics[width=\linewidth]{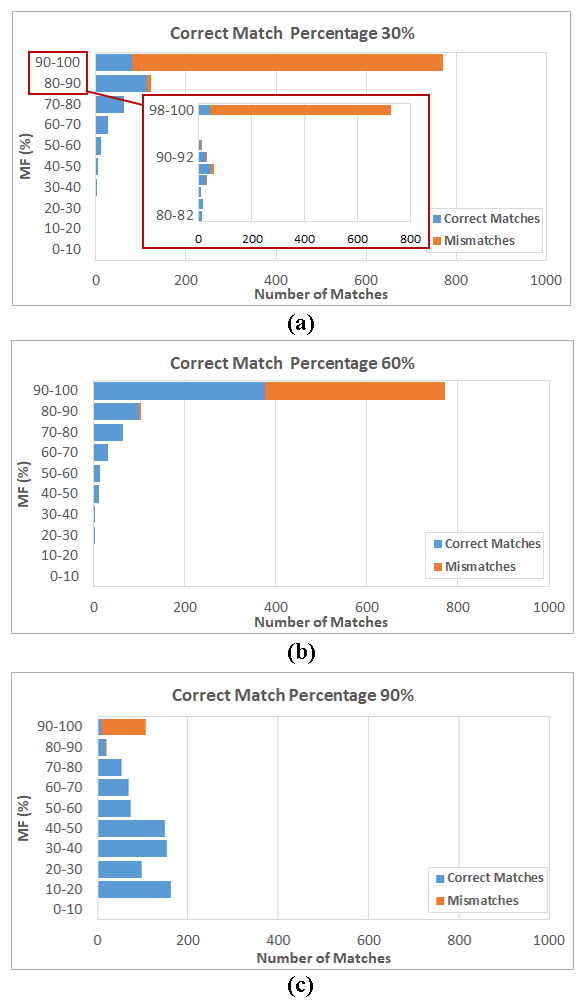}
    \caption{Histogram of $MF$ values for three sample synthetic data experiments with 1000 matches and 30\%, 60\% and 90\% of correct matches.}
    \label{outlier_removal_algorithm_1}
\end{figure}

\begin{figure*} 
    \centering
    \includegraphics[width=\linewidth]{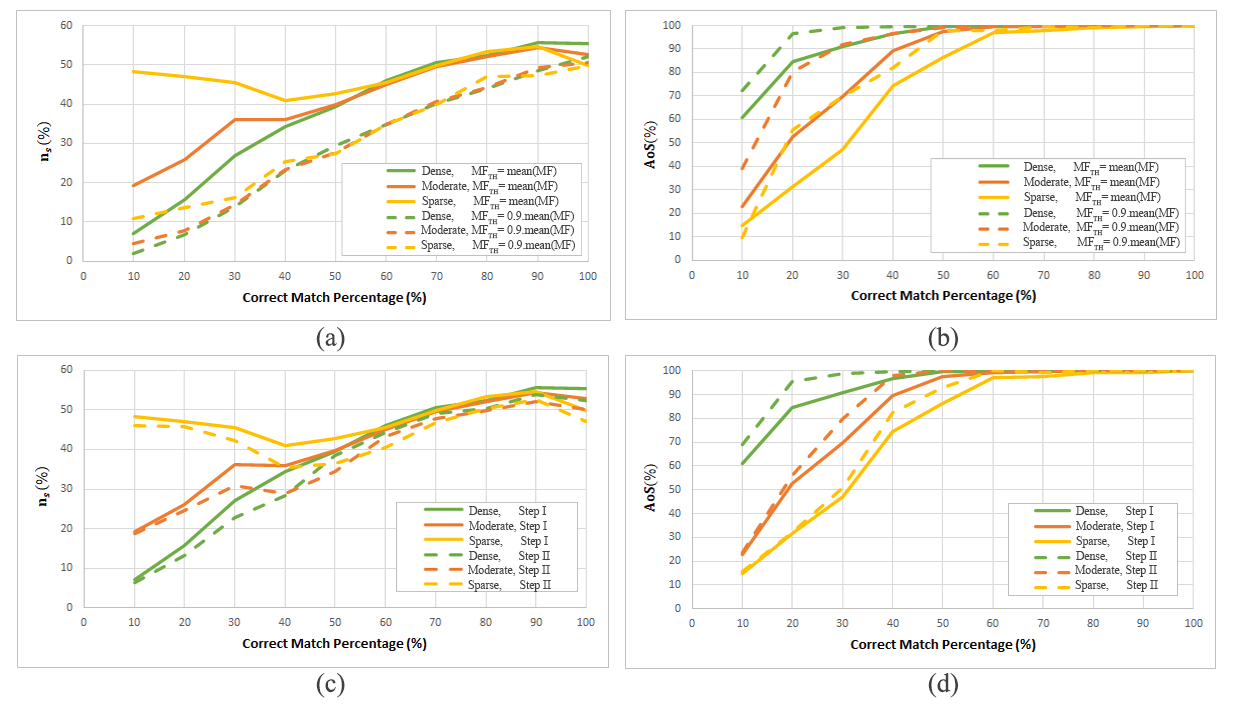}
    \caption{Results of applying the first two steps of the algorithm $\tt{myNeighbor}$ in synthetic data experiments in three different scenarios; Dense (1000 matches), Moderate (200 matches), and Sparse (50 matches). Each curve is the average result of 1000 trials. The first row gives $n_s$ and $AoS$ from \textit{Step I} for two different values of $MF_{TH}$. The second row gives the results of \textit{Step II} in comparison to the results of \textit{Step I} with $MF_{th}=\text{mean}(MF)$.}
    \label{outlier_removal_data}
\end{figure*}

\subsection{Synthetic data experiments for calibrating parameters}  \label{subsec:Synthetic Data Experiments For Calibrating Parameters}

These experiments are conducted by synthetically forming two images of a mesh ${M_T}$ and a series of matches between the two images.
The first image shows ${M_T}$ in its flat rest shape with all its keypoints on it.
We call this image $\mathcal{I}_{F}$.
In $\mathcal{I}_{F}$, the keypoints can be considered as the extracted keypoints from $\mathcal{P}$ and the 2D mesh is equivalent to ${M}$. 
The second image simulates $\mathcal{I}$ and shows ${M_T}$ having undergone a random 3D deformation. We call this deformed mesh ${M_G}$.
The keypoints in this image can be positioned in their correct locations on the mesh (correct matches) or being displaced in the image area (mismatches).

We consider ${M_T}$ as a regular triangular mesh with $10\times6$ points in 3D space.
In order to deform ${M_T}$, we use the same method as in~\cite{arandamonocular}. 
This is done by applying two 3D deformations containing random translations and rotations to two mesh cells at both sides of ${M_T}$. 
The deformation is calculated in an iterative process based on position-based dynamics~\cite{muller2007position,bender2014survey}.
As for generating keypoints, we first randomly place keypoints in the inner area of ${M}$ in $\mathcal{I}_{F}$. In order to create the matches between $\mathcal{I}_{F}$ and $\mathcal{I}$, we then transfer the keypoints from $\mathcal{I}_{F}$ to $\mathcal{I}$ using a three-step process: calculating barycentric coordinates of the keypoints in ${M}$, transferring the keypoints to the 3D deformed mesh using the  barycentric coordinates and the new 3D mesh points of the deformed ${M_T}$, and eventually projecting the transferred keypoints to $\mathcal{I}$. To generate mismatches, an arbitrary percentage of the transferred keypoints were corrupted by randomly distributing them all over the area of $\mathcal{I}$. Two samples of the generated images for 100 and 1000 matches each with 30\% mismatches can be observed in the two first columns of Figure~\ref{image_outlier_removal_syn}.

\subsection{Methodology}  \label{subsec:Methodology}

The algorithm $\tt{myNeighbor}$ is applied on $N_m$ matches denoted as $C_p \leftrightarrow C_q$ between $\mathcal{P}$ and $\mathcal{I}$, with:
\begin{equation}
C_p = \{p_{1}, ..., p_{N_m}\} ,\;\;p_{i}=(x_i,y_i)
\end{equation}
\begin{equation}
C_q = \{q_{1}, ..., q_{N_m}\} ,\;\;q_{i}=(u_i,v_i)
\end{equation}
A pair $(p_i,q_i)$ of points with the same index forms a match $p_i \leftrightarrow q_i$. 
We define the set of correct matches $S_{in}$ as the collection of matches $p_i \leftrightarrow q_i$ where $p_i$ and $q_i$ point to the same location on the deforming surface in $\mathcal{P}$ and $\mathcal{I}$. On the contrary, when the pointing locations of the match points are different, they are categorized as mismatches $S_{out}$. 
The goal of $\tt{myNeighbor}$ is to form and remove the subsets $O_p\subset C_p$ and $O_q\subset C_q$ which have the largest possible number of matches belonging to $S_{out}$ and smallest possible number of matches belonging to $S_{in}$. 
We explain the steps of our algorithm to fulfill this goal.

\subsubsection{Step I – Neighbor-based correct match selection}
We select subsets $C_{p_s}\subset C_p$ and $C_{q_s}\subset C_q$ which are highly probable to form correct matches.
We start by defining $W_G$ as the groundtruth warp between $\mathcal{P}$ and $\mathcal{I}$ that can transfer all the match points $C_p$ from $\mathcal{P}$ to their correct locations in $\mathcal{I}$. With this definition, we have the set of correct matches $S_{in}$ as:
\begin{equation}
{S_{in}} =  \{(p_i,q_i) \,|\, i\in R \},
\end{equation}
where:
\begin{equation}
R = \{i \,|\,  \| W_G(p_i)-q_i \| < \epsilon \},
\end{equation}
where $\epsilon$ is a very small positive number. 
Warp $W_G$ is an unknown composition of isometric deformation and perspective projection mappings. The isometric deformation mapping preserves the geodesic distances among the points and their topological structure on the object's surface. However, with the addition of perspective projection mappings, only the topological structure of points remains preserved in visible areas. This implies that by applying $W_G$, the neighborhood structure among the points on the object in $\mathcal{P}$ and $\mathcal{I}$ should be preserved.
We exploit this characteristic of $W_G$ to estimate $\widehat{R}$ as the set of indices of highly probable correct matches $C_{p_s} \leftrightarrow C_{q_s}$.
 To do so, first, we form two Delaunay triangulations, $T_{p}=D(C_p)$ in $\mathcal{P}$, and $T_{q}=D(C_q)$ in $\mathcal{I}$. Then, for each match $i$, we calculate two sets of first-order neighbors ${Q_p}(i)$ and ${Q_q}(i)$ in $\mathcal{P}$ and $\mathcal{I}$, respectively. We then define the \textrm{\textit{Mismatch Factor}} ($MF$) criterion for match $i$ as:
\begin{equation} 
MF(i) = \frac{|{Q_p}(i)\cup {Q_q}(i) - {Q_p}(i)\cap {Q_q}(i)|}{|{Q_p}(i)\cup {Q_q}(i)|}\times100
\end{equation}
For each match, $MF$ represents the difference in the neighbor points between $\mathcal{P}$ and $\mathcal{I}$ as a percentage. 
Ideally, we expect that for all the matches $MF=0$, which implies that there is no difference in the neighbors of each match during a deformation.
However, in practice, there are two reasons which rather put $MF$ values in a range from 0 to 100: the presence of mismatches and variations in triangulation. The presence of mismatches can affect the value of $MF$ in two ways. First, when the match point $i$ in $\mathcal{I}$ is a mismatch and thus located in a wrong location. And second, when the match point $i$ in $\mathcal{I}$ is a correct match but one, several, or all of its neighbors are mismatches. Both of these cases result in different neighbors in $\mathcal{I}$ in comparison to $\mathcal{P}$. 
As for the two triangulations, it should be noted that even in the absence of mismatches, the neighborhood structures in $T_{p}$ and $T_{q}$ do not necessarily coincide. This is because of surface deformation, change in viewpoint, and occlusions.

Calculating $MF$ for all the matches, we can have a fair estimation regarding the state of the matches. 
The lower values of $MF(i)$ indicate that the match $i$ is surrounded by similar matches in $\mathcal{P}$ and $\mathcal{I}$ and has a higher probability to be placed in its correct location and thus be a correct match. 
On the contrary, the higher values of $MF(i)$ can stem from the wrong location of the match $i$ in comparison to its neighbors which strengthens the possibility of it being a mismatch. 
The basic idea in this step is to form $C_{p_s} \leftrightarrow C_{q_s}$ by selecting pairs of highly probable correct matches $p_s \leftrightarrow q_s$. This is done by choosing the matches with lower values of $MF$. 
We examined the validity of this reasoning by evaluating three different synthetic data experiments, each with 1000 matches and different rates of correct matches (30\%, 60\%, and 90\%). Figure~\ref{outlier_removal_algorithm_1} shows the histogram of $MF$ for each case.
We observe that the dispersion of $MF$ spans a wider range as the value of the correct match rate grows. 
For higher numbers of correct matches, there are more similarities in the neighbor lists of each match and, consequently, $MF$ decreases.
Furthermore, regardless of the values of the correct match rate, the majority of the mismatches are accumulated in the top bins of the graphs that correspond to higher values of $MF$. This is shown in more detail for the case with the correct match percentage of 30\% by expanding the last two bins of the graph in Figure~\ref{outlier_removal_algorithm_1}.a. 
This validates our prior reasoning that by selecting the matches with $MF$ below a certain threshold $MF_{th}$, we can have a set of matches which are highly probable to be correct. 
To quantify the appropriateness of this selection, we define two criteria, based on the following two quantities. 
The first quantity is $n_s$, which is the percentage of the selected matches compared to the total number of matches:
\begin{equation}
n_s = \frac{ \left| C_s \right|  }{ N_m }\times 100,
\end{equation}
where $C_s=\{ (p_{i},q_{i})\,|\,i\in \widehat{R} \}$ is the set of selected matches.
The second quantity is $AoS$, which is the Accuracy of Selection, defined as:
\begin{equation}
AoS = \frac{ \left| C_s \cap S_{in} \right|}{ \left| C_s \right| }\times 100.
\end{equation}
Our goal is to choose the value of $MF_{th}$ in the way that we have both of these criteria to be as high as possible, which means selecting a high percentage of matches with high accuracy. However, practically, these two criteria work in reverse. By choosing a higher value for $MF_{th}$, more matches are selected (higher $n_s$) but with less accuracy (lower $AoS$) and vice versa. 
In order to choose the proper value for $MF_{th}$, we analyzed the behavior of these two criteria for a series of synthetic data experiments.
\begin{figure*}
    \centering
    \includegraphics[width=\linewidth]{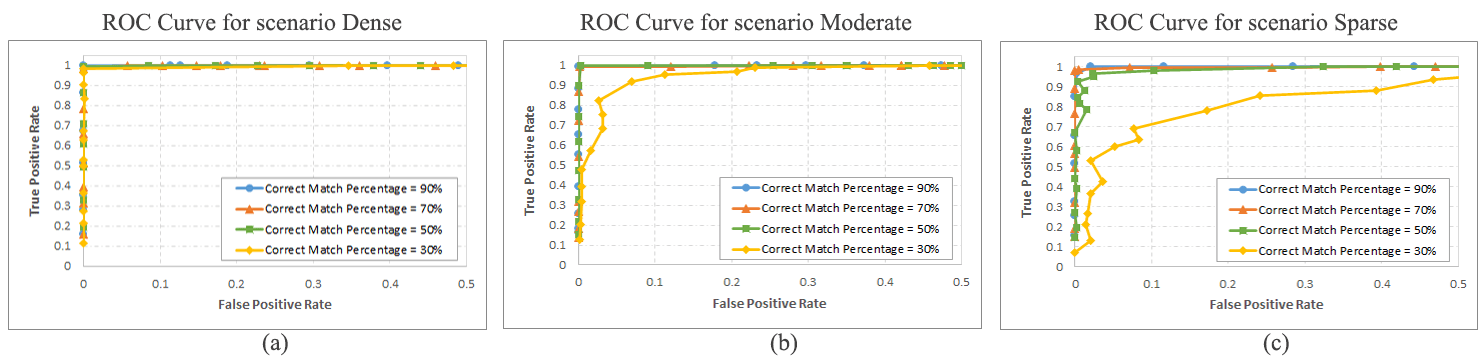}
    \caption{ROC curves resulting from the algorithm $\tt{myNeighbor}$ in synthetic data experiments in three scenarios; Dense (1000 matches), Moderate (200 matches), and Sparse (50 matches). Each point is the average result of 1000 trials calculated with a specific value of $d_{3_{th}}$.}
    \label{outlier_removal_data2}
\end{figure*}
\begin{figure*}
    \centering
    \includegraphics[width=\linewidth]{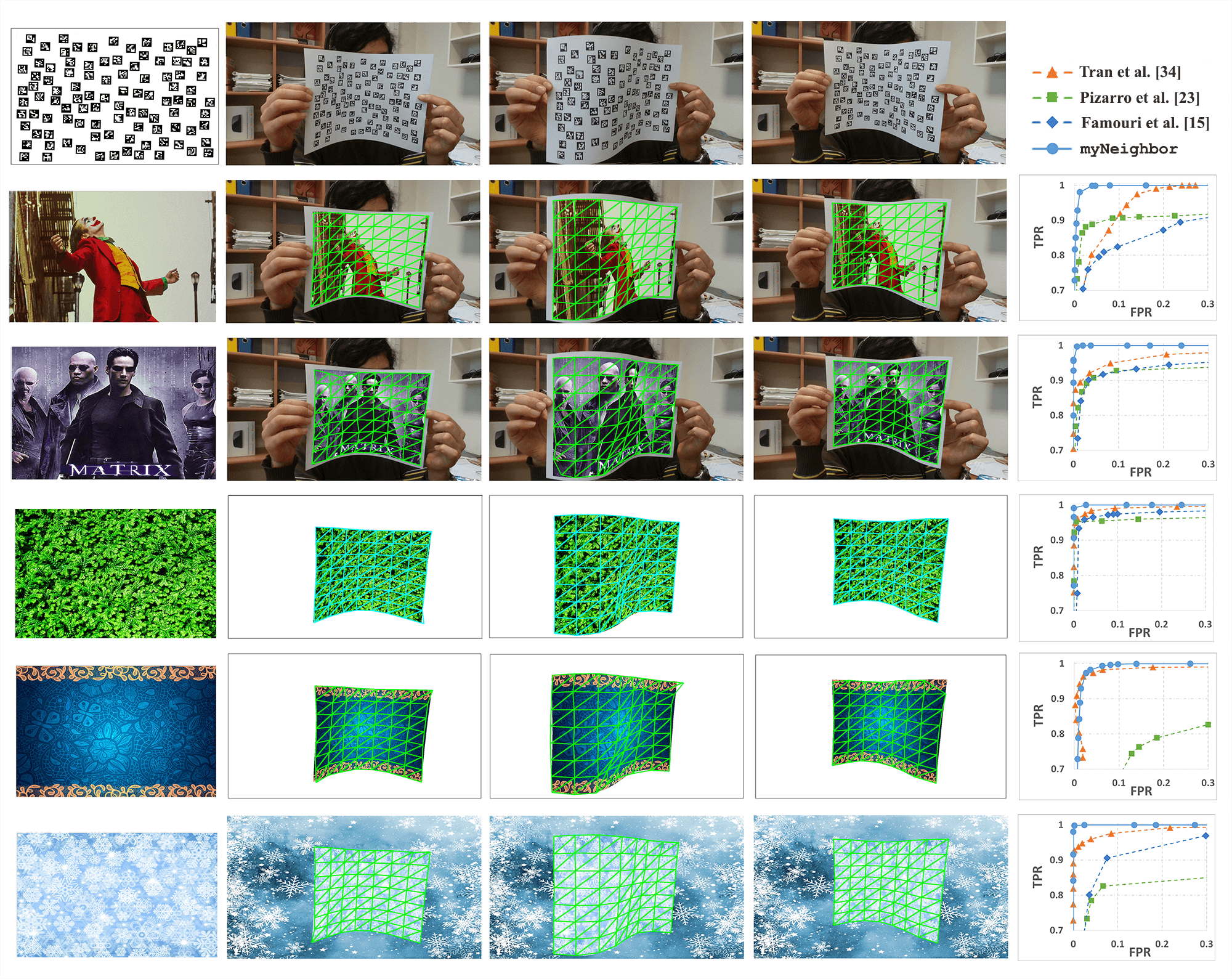}
    \caption{Performance evaluation of our mismatch removal method $\tt{myNeighbor}$ in comparison to the state-of-the-art methods using the $\tt{FREX}$ protocol. The first row shows the Aruco template and three selected images (14, 47, 60) of the deformation of the printed Aruco template. The following rows show five datasets of generated scenes with the texturemap in the first column, three generated images corresponding to the first row in the next columns, and the ROC curves of the mismatch removal algorithms in the last column. For each of the images $\widehat{M}_3$ from $\tt{myNeighbor}$ is overlaid.}
    \label{image_outlier_removal_experimentWarp}
\end{figure*}
We consider three scenarios for these experiments based on the number of matches, i.e., Dense, Moderate, and Sparse with in turn 1000, 200, and 50 total number of matches.
The experiments were done in a wide range of correct match percentages (10\% to 100\%) for each scenario. 
Two different values of the criterion $MF_{th}$ were studied; $mean$ and $0.9\times mean$ where $mean$ is the mean of all $MF$ values in each experiment.
The results are presented in Figure~\ref{outlier_removal_data}.a and~\ref{outlier_removal_data}.b.
Each point in the graph is the average result of 1000 trials.
The first point that should be noted here is that, generally, the proposed match selection method in this step is more reliable as the number of total matches grows. This can be deduced by comparing the higher values of $AoS$ in the Dense case with the ones in the Moderate and Sparse cases.
As for choosing $MF_{th}$, it should be noted that setting $MF_{th}=0.9\times mean$ leads to higher values of $AoS$ in comparison to the case with $MF_{th}=mean$. Nevertheless, as shown in Figure~\ref{outlier_removal_data}.a, this sacrifices a high percentage of matches by dropping $n_s$ significantly, which is undesirable. Hence, in this step, we choose $mean$ as the value of $MF_{th}$ and form $\widehat{R}$ as the set of indices of probable correct matches. While this choice implies a higher number of selected mismatches (lower $AoS$), we note that these mismatches can be removed in \textit{Step II}.

As the final operation in this step, we estimate the warp $W_1$ between $\mathcal{P}$ and $\mathcal{I}$ using the selected matches $C_{p_s} \leftrightarrow C_{q_s}$. We then exploit this warp to transfer $M$ to $\mathcal{I}$. We call this new mesh ${\widehat{M}_1}$. 
As can be seen in the third column of Figure~\ref{image_outlier_removal_syn}, the mesh ${\widehat{M}_1}$ (shown in orange) may not be totally faithful to the deformation of ${M_G}$ in $\mathcal{I}$, which is due to the inaccuracies in the calculation of the warp $W_1$. This stems from two main reasons; the existence of mismatches in our selection (shown as red dots), and the insufficient number of correct matches in some areas. 
In the next step, we exploit the transferred mesh ${\widehat{M}_1}$ to remove the possible remaining mismatches from the selected matches. 

\begin{table}[htb]
    \centering
    \scalebox{1.2}{%
    \begin{tabular}{cc}
    \toprule
        Method & \shortstack{Average run-time (s)} \\
    \midrule
        $\tt{myNeighbor}$ & 0.0139 \\
        Tran et al.~\cite{tran2012defence} & 0.0206\\
        Pizarro et al.~\cite{pizarro2012feature} & 1.8925\\
        Famouri et al.~\cite{famouri2018fast} & 0.0171\\
    \bottomrule
    \end{tabular}}
    \caption{Comparison of the average run-time of the mismatch removal algorithms for processing all the images of all the datasets.}
    \vspace{3mm}
    \label{outlier_removal_time_comparison}
\end{table}
\begin{figure*}
    \centering
    \includegraphics[width=\linewidth]{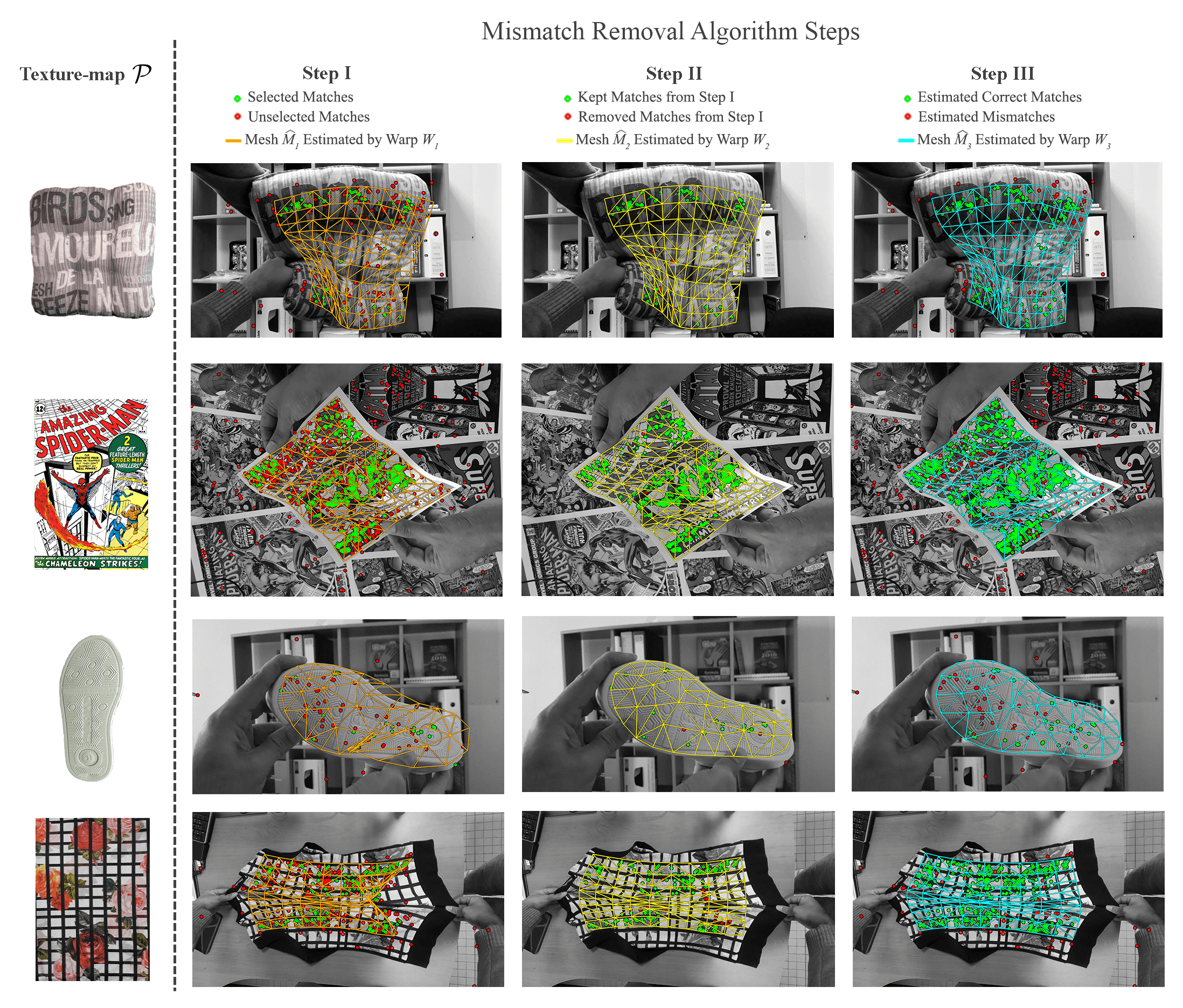}
    \caption{Applying $\tt{myNeighbor}$ on four real cases: a cushion, a Spiderman poster, a shoe sole, and an elastic shirt. The first column shows the texturemaps. The second column shows \textit{Step I}. All the matches are shown in this column while the selected matches in \textit{Step I} are shown in green. These selected matches are transferred to column three that shows \textit{Step II}. In this column, those matches which are chosen as possible mismatches are shown in red. The last column is the distinction between the estimated correct matches (in green) and the estimated mismatches (in red) in \textit{Step III}. The meshes $\widehat{M}_1$, $\widehat{M}_2$, and $\widehat{M}_3$ are overlaid to illustrate the computed warps.  }
    \label{image_outlier_removal_real}
\end{figure*}
\begin{figure*}
    \centering
    \includegraphics[width=\linewidth]{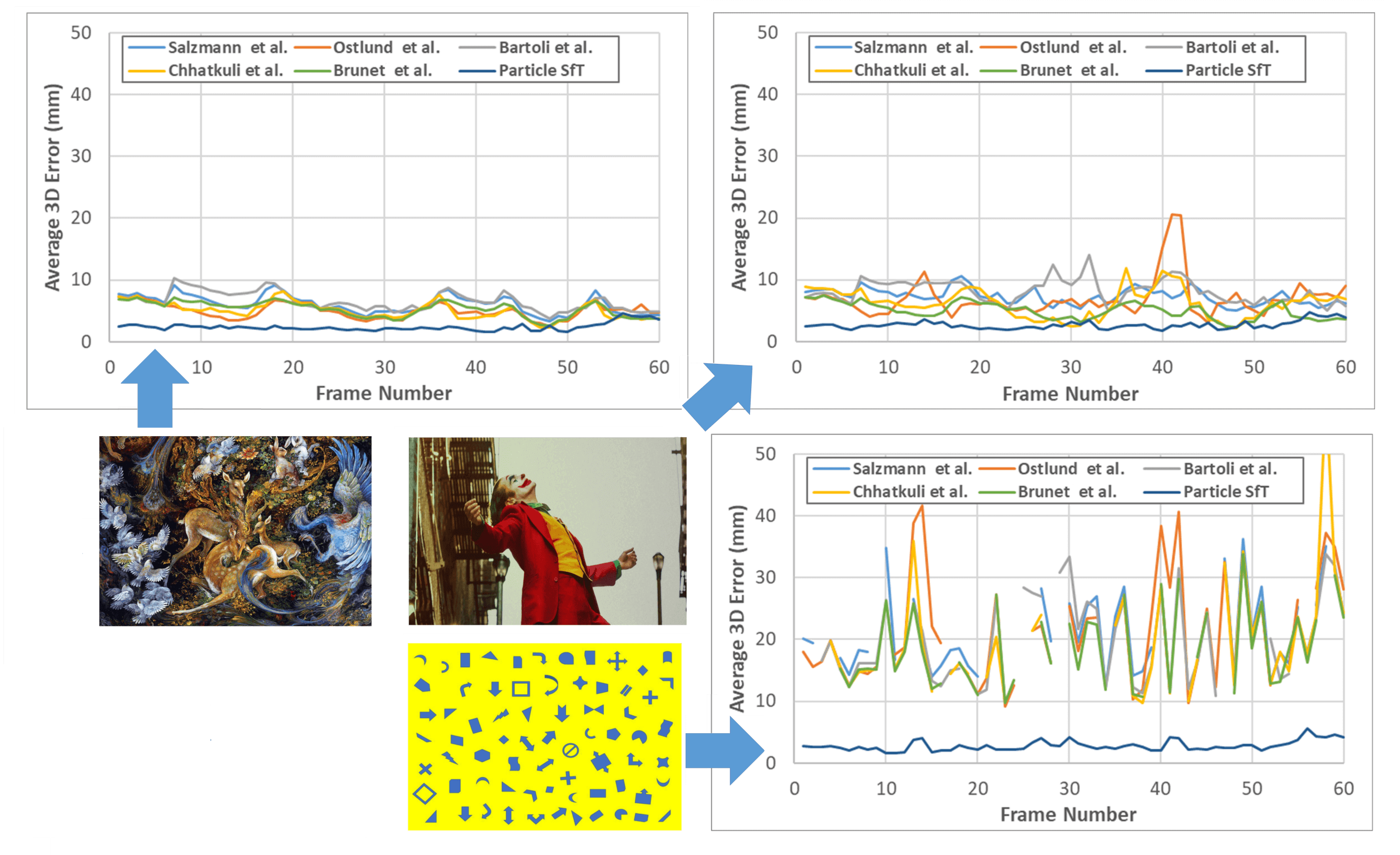}
    \caption{Comparing the accuracy of the 3D shape inference methods with Particle-SfT with three datasets obtained by $\tt{FREX}$. The 3D shape inference methods are Brunet et al.~\cite{brunet2014monocular}, Chhatkuli et al.~\cite{chhatkuli2014stable}, Bartoli et al.~\cite{bartoli2015shape}, Ostlund et al.~\cite{ostlund2012laplacian}, and Salzmann et al.~\cite{salzmann2010linear}.}
    \label{image_3dReconstruction_fre}
\end{figure*}


\begin{figure}
    \centering
    \includegraphics[width=\linewidth]{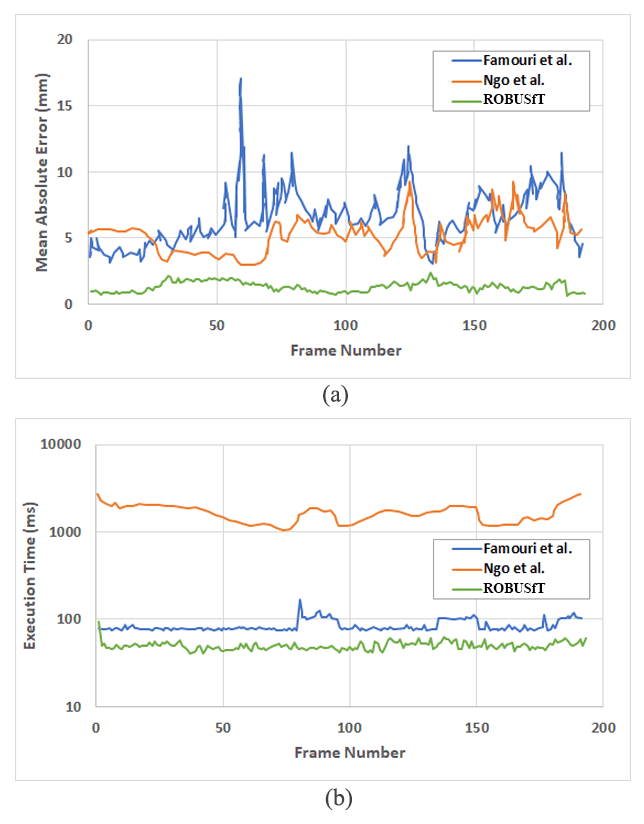}
    \caption{Comparison between $\tt{ROBUSfT}$ and the methods presented by Famouri et al.~\cite{famouri2018fast} and Ngo et al~\cite{ngo2015template} on the public dataset provided in~\cite{varol2012monocular}. (a) Mean absolute 3D error between the inferred shape and the groundtruth. (b) Execution time in milliseconds.}
    \label{image_3d_comparison}
\end{figure}

\begin{figure*}
    \centering
    \includegraphics[width=\linewidth]{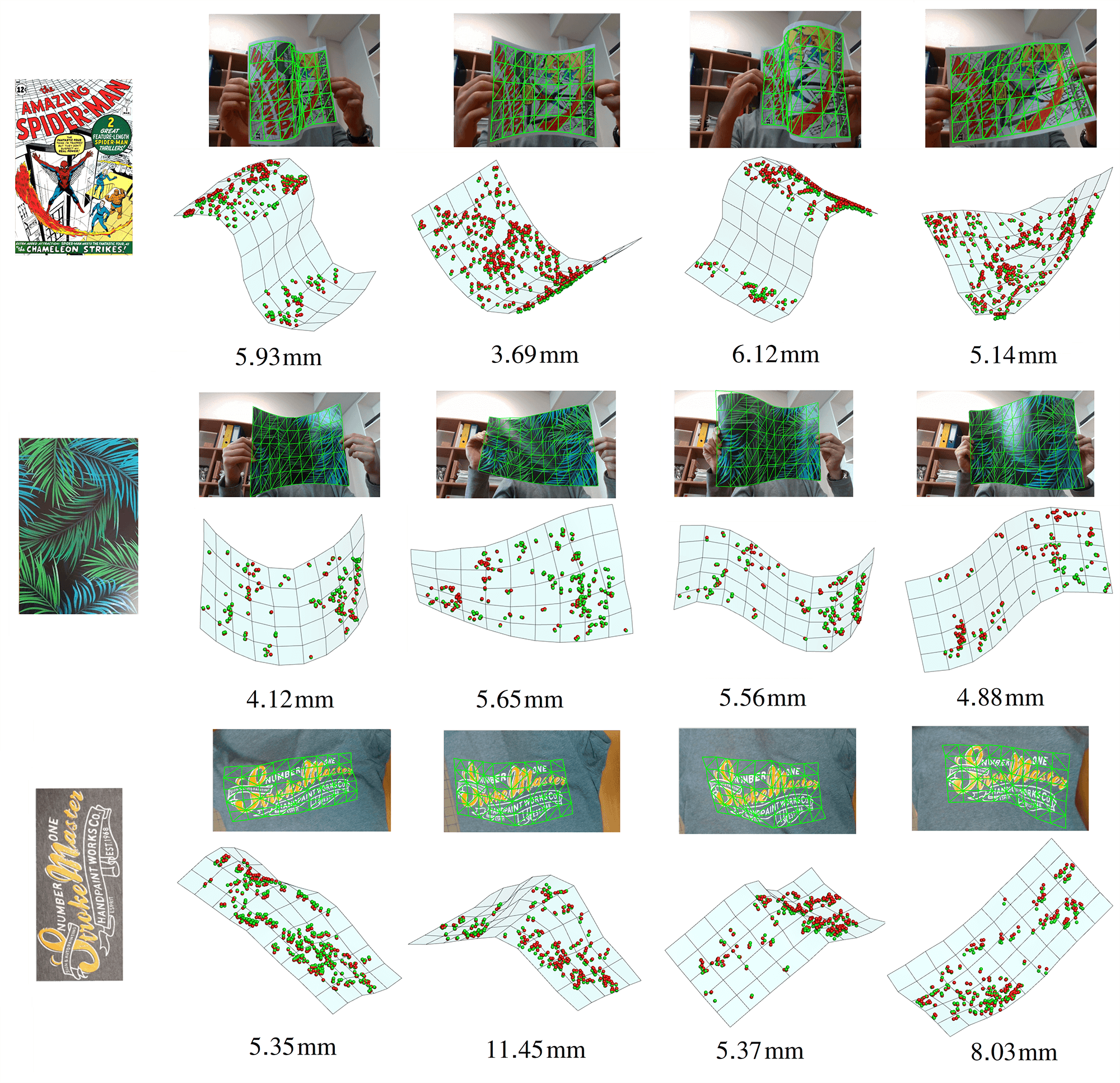}
    \caption{Evaluating $\tt{ROBUSfT}$ in three real data experiments; a Spiderman  poster, a chopping mat, and a t-shirt. The texturemaps of the templates are shown in the first column. For each case, four images are shown. Below each frame, the reconstructed 3D shape of the deforming object with the estimated 3D coordinates of the estimated correct matches (red particles) as well as their ground truth (green particles) are shown. The 2D projections of the 3D inferred shapes are also overlaid on the image. For each image, the median Euclidean distance between the estimated 3D coordinates of the estimated correct matches and their ground truth is given below the reconstructed shape. }
    \label{real_experiment}
\end{figure*}

\begin{figure}
    \centering
    \includegraphics[width=\linewidth]{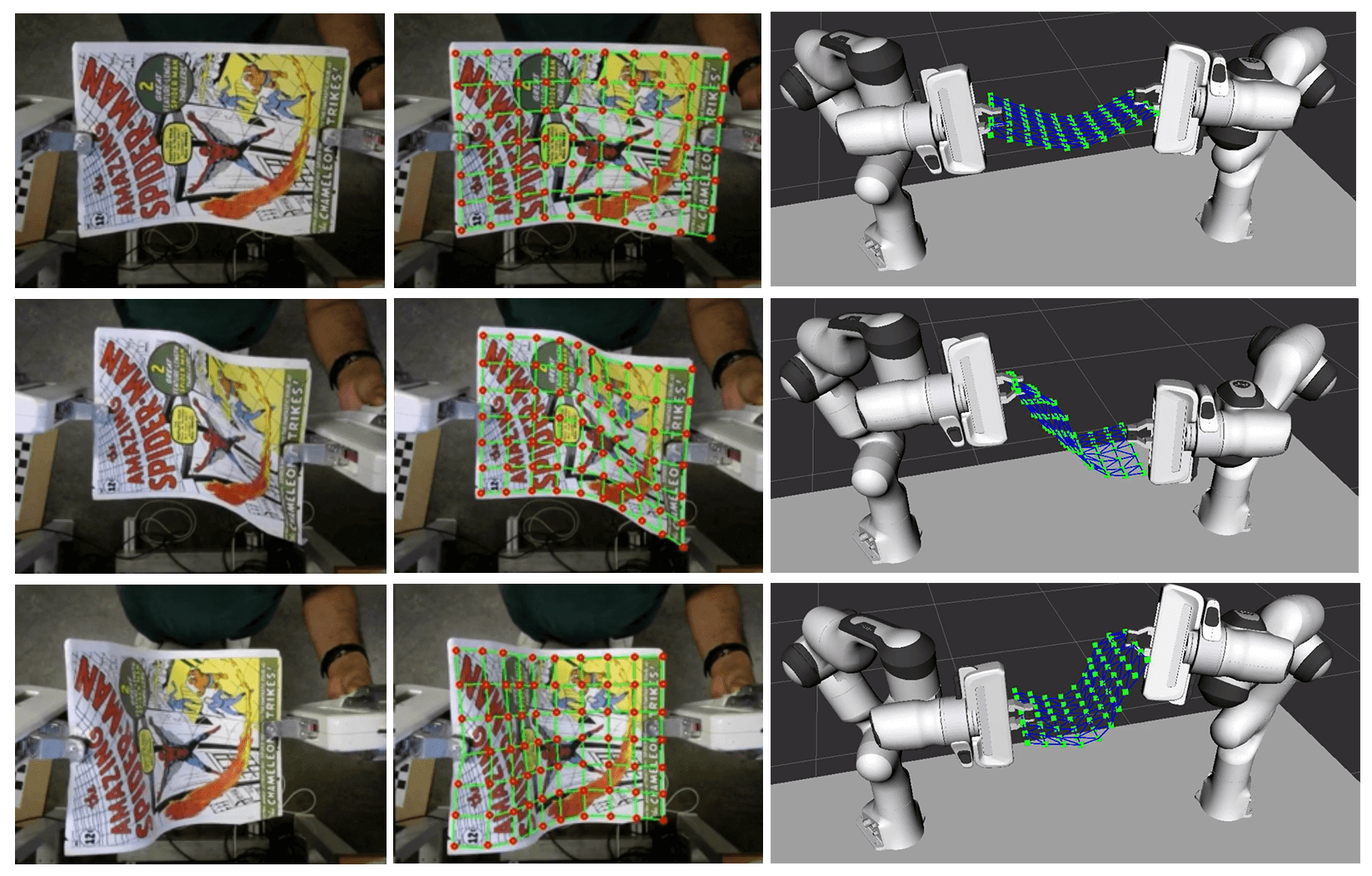}
    \caption{Evaluating $\tt{ROBUSfT}$ in a real data experiment with two robotic arms; soft constraints are applied to bind the constrained mesh points to the grippers. Each row shows three images: the original camera view, the projection of the 3D reconstructed mesh on the camera view, and the 3D reconstructed mesh with the robots in the RViz environment.}
    \label{robot_experiment}
\end{figure}
\subsubsection{Step II – Removing mismatches from the list of selected matches} 
We remove the possible mismatches from the selected matches $C_{p_s} \leftrightarrow C_{q_s}$. We first form the set $C_{\hat{q}_s}$ by transferring $C_{p_s}$ to $\mathcal{I}$.
This is done by finding the barycentric coordinates of each selected match $p_{s_i}\in C_{p_s}$ with respect to ${M}$ and applying them on the transferred 2D mesh $\widehat{M}_1$ from \textit{Step I}. 
We then use the following decision criterion to identify and remove possible mismatches one by one from the selected matches $C_{p_s}\leftrightarrow C_{q_s}$:
\begin{equation}
\Big| {d_2}(i) - \text{median}\big(\{{d_2}(j)\}\big) \Big|   \geqslant 2.5 \, \text{MAD}, \label{eq:mad}
\end{equation}
where ${d_2}(i) = \| \hat{q}_{s_i} - q_{s_i}\|$ with $i\in \widehat{R}$. 
MAD (Median of Absolute Deviations from Median) is calculated as:
\begin{equation}
\text{MAD} = k\,\,\text{median}\Big(\Big\{\Big| {d_2}(i) - \text{median}\big(\{d_2(j)\}\big) \Big|\Big\}\Big),
\end{equation}
where $k=1.4826$ is a constant number. 
The values of ${d_2}$ are relatively larger for mismatches in comparison to correct matches. 
This stems from two reasons. First, the small percentage of mismatches compared to the great majority of correct matches coming from \textit{Step I} and thus lesser influence of mismatches in the estimation of warp $W_1$. Second, the inconsistent location of mismatches in $\mathcal{P}$ and $\mathcal{I}$.
The decision criterion in equation \eqref{eq:mad} is chosen due to the distribution type of ${d_2}$, with the presence of just a small percentage of large values among the majority of small values.
Figure~\ref{outlier_removal_data}.c and d illustrate the result of this step. 
As can be seen, unlike the previous strategy of choosing a smaller $MF_{th}$, this method results in improvement of $AoS$ without losing a considerable percentage of selected matches. 
This can be clearly observed by comparing $n_s$ in Figures~\ref{outlier_removal_data}.a and c.

As the last operation in this step, warp $W_2$ is calculated using the purified selected matches $C_{p_s} \leftrightarrow C_{q_s}$. 
This warp is then used to transfer $M$ to the image space and form $\widehat{M}_2$. 
The result of removing possible mismatches in this step along with the transferred mesh $\widehat{M}_2$ are shown in the fourth column of Figure~\ref{image_outlier_removal_syn}. 
As can be observed, in comparison to $\widehat{M}_1$, $\widehat{M}_2$ has a better compliance to $M_G$. 


\subsubsection{Step III – Extracting mismatches from the list of all the matches}
In this step, we exploit the transferred mesh $\widehat{M}_2$ to extract the mismatches $O_{p} \leftrightarrow O_{q}$ from the total matches $C_{p} \leftrightarrow C_{q}$. The process is similar to \textit{Step II} except that this time all of the matches are checked. 
We first transfer the template match points $C_p$ to the image space and form the set $C_{\hat{q}}$. This is done by calculating barycentric coordinates of all the match points $C_{p}$ with respect to $M$ and applying them on the new transferred mesh $\widehat{M}_2$. 
We define the following decision criterion to detect and remove mismatches:
\begin{equation}
{d_3}(i) = \| \hat{q}_i - q_i\| \geqslant d_{3_{th}}
\end{equation}
Unlike \textit{Step II} where we used the MAD criterion to remove just a small rate of mismatches, this time we use a constant threshold $d_{3_{th}}$. This is due to the higher percentage of mismatches compared to \textit{Step II}.
In order to make this distinction method more robust, we consider $d_{3_{th}}$ as the multiplication of a sample length ${l_s}$ and a constant coefficient ${\alpha_s}$. The sample length ${l_s}$ is a measure of the size of the object in the image in pixels and is calculated as the average distance between all the mesh points in the transferred mesh $\widehat{M}_2$. 
To choose a proper value for the constant coefficient ${\alpha_s}$, a series of synthetic data experiments with the same three scenarios as before (Dense, Moderate, and Sparse) and four different correct match rates was performed. The results are presented as ROC (Receiver Operating Characteristic) curves in Figure~\ref{outlier_removal_data2}.a-c. Each point represents the average TPR (True Positive Rate) versus the average FPR (False Positive Rate) computed in 1000 trials using a specific value of ${\alpha_s}$ in the range of $[0,1]$.
TPR is calculated as the number of selected true mismatches over the number of all true mismatches, and FPR is calculated as the number of true correct matches mistakenly selected as mismatches over the number of all true correct matches.
Ideally, all the mismatches should be discarded (TPR=100\%) without discarding any correct matches (FPR=0\%). 
Hence, the most favorable ${\alpha_s}$ in a single ROC curve is the one that results in the maximum possible TPR leaving the FPR below a reasonable value. 
We choose ${\alpha_s} = 0.15$ which keeps TPR above 90\% while FPR remains below 10\% for most of the cases. 
The last column of Figure~\ref{image_outlier_removal_syn} illustrates the estimated correct matches (in green) and the estimated mismatches (in red) for each case. 
We also use the estimated correct matches to estimate warp $W_3$ and transfer $M$ to $\mathcal{I}$ and form $\widehat{M}_3$ (shown in cyan). As can be seen, there is a high compliance between $\widehat{M}_3$ and $M_G$.
It should be noted that estimating $W_3$ and $\widehat{M}_3$ is not necessary in $\tt{myNeighbor}$ and we merely estimate them just to visually present the effectiveness of the algorithm in removing the mismatches. However, considering $\tt{myNeighbor}$ as a step in $\tt{ROBUSfT}$, due to the fact that the final estimated correct matches are passed from this step to \textit{Step 3} of $\tt{ROBUSfT}$ which is warping, $W_3$ and $\widehat{M}_3$ can also represent ${W}$ and $\widehat{M}$ in the warping step, respectively.

\subsection{Mismatch removal results}  \label{subsec:Mismatch removal results}
In this section, we demonstrate the efficiency of $\tt{myNeighbor}$ by evaluating its performance through various tests. 
We first compare the results of the algorithm with the state-of-the-art algorithms in the literature by testing them through $\tt{FREX}$. The experiment includes 60 frames of continuous deformation of the Aruco template in front of the camera. 
Five datasets were generated in this experiment each with an arbitrary texture with a challenging pattern. Three different types of backgrounds were also considered for these five cases, specifically two original backgrounds, two white backgrounds, and a background with a pattern similar to one of the texturemaps. 
We apply all the mismatch removal algorithms on all datasets. 
For each dataset, the corresponding arbitrary texture was used as the texturemap for the mismatch removal algorithms. 
The matches between the texturemap and each image of the dataset are extracted using SIFT.
The results are presented in Figure~\ref{image_outlier_removal_experimentWarp}. 
The first row illustrates the Aruco template and also three selected original images of its deformation in front of the camera. 
The lower rows represent the five datasets generated by $\tt{FREX}$. Each row shows the arbitrary texture of the dataset in the first column, the three selected generated images, and eventually the resulting ROC curves for all the mismatch removal algorithms on the dataset. In the ROC curves, for a certain algorithm and a certain dataset, each point is the average value of TPR and FPR over all 60 images of that dataset using a specific value for the threshold used in the algorithm.
As can be seen, in all cases, our algorithm outperforms the other algorithms. 
In order to show the performance of our algorithm visually, for each dataset, we overlaid $\widehat{M}_3$ for the three selected frames. 
As can be observed, the transferred meshes are visually well-aligned to the 2D deformed shape of the object. 
In some cases, a small number of irregularities can be observed in certain areas (for example in the Matrix poster). This is because of the presence of a small number of mismatches in our list of estimated correct matches and the lack of matches in those areas.
As for comparing the execution speed of different mismatch removal algorithms, the process run-times for all the frames of all datasets were averaged and tabulated in Table~\ref{outlier_removal_time_comparison}. It shows that our algorithm is faster than the others. It should be however noted that our algorithm is implemented in C++ while the others are in Matlab. 

After validating the efficiency of $\tt{myNeighbor}$ in comparison to the state-of-the-art algorithms in the literature, we evaluate its performance in real cases. 
To this end, we applied our algorithm to four real deforming objects as shown in Figure~\ref{image_outlier_removal_real}. 
We chose these cases in such a way that each one is challenging in a special way.
The cases include a cushion with non-smooth surface and severe deformation, a Spiderman poster deformed in a scene with background covered with almost the same posters, a shoe sole with an almost repetitive texture, and a shirt with elastic deformation. 
The texturemaps are shown in the first column of Figure~\ref{image_outlier_removal_real}. 
The second to fourth columns show the results of \textit{Step I} to \textit{Step III} of $\tt{myNeighbor}$.
In each step, the alignment of the corresponding transferred mesh to the 2D shape of the deforming object can be considered as an indication of the correctness and abundance of the estimated correct matches. Like in the synthetic data experiments, this alignment improves progressively in different steps of our algorithm.
One point that should be noted here is that the shirt (the last case in Figure~\ref{image_outlier_removal_real}) is elastic. We exert a non-isometric deformation on it by pulling from both sides, and $\tt{myNeighbor}$ still works. 
This is due to the fact that we did not make any assumption regarding isometry. In fact, the only assumption that we made is the preservation of neighborhood structure in the deforming object. As a result, $\tt{myNeighbor}$ also works with non-isometric deformations which preserve neighborhood structure.

\section{Experimental Results} \label{sec:experiments}
We evaluate the performance of $\tt{ROBUSfT}$ on different deforming objects in various conditions. We divide this section into two main parts; first, comparing the results with the state-of-the-art methods and then evaluating $\tt{ROBUSfT}$ in several other challenging cases.

\subsection{Comparison to the state-of-the-art methods} \label{subsec:Comparison-to-state-of-the-art-methods}
We compare $\tt{ROBUSfT}$ with the state-of-the-art methods through two different tests. 
The first test is conducted among the shape inference methods (\texttt{G1}). The second test is carried out among the integrated methods (\texttt{G2}). 

We use $\tt{FREX}$ to conduct the first test. To this end, the same 60 images of deforming Aruco marker paper sheet are used. 
We create three different datasets using three arbitrary texturemaps and apply a white background to all the scenes.
The arbitrary texturemaps include a painting, the Joker poster, and a paper sheet filled with basic geometric shapes.
These images are shown in Figure~\ref{image_3dReconstruction_fre}.
In each dataset, we compare the result of the last two steps of $\tt{ROBUSfT}$ (warp estimation and 3D shape inference) with five other shape inference methods from Brunet et al.~\cite{brunet2014monocular}, Chhatkuli et al.~\cite{chhatkuli2014stable}, Bartoli et al.~\cite{bartoli2015shape}, Ostlund et al.~\cite{ostlund2012laplacian}, and Salzmann et al.~\cite{salzmann2010linear}.
A similar comparison was made in~\cite{ozgur2017particle} on another dataset. However, in~\cite{ozgur2017particle}, a random 3D shape was used as the initial guess for Particle-SfT algorithm in each image of the video; in contrast, we use the 3D inferred shape of the object in each image as the initial guess for the next image. 
In each dataset, the matches between $\mathcal{P}$ and each image are extracted using SIFT. We then separate the correct matches and use them as the input for all the methods. 
If required by a shape inference method, a BBS warp is estimated based on these correct matches and used as the input to that shape inference method.
The results for all three datasets are presented in Figure~\ref{image_3dReconstruction_fre} as the average 3D error between the 3D inferred shapes and the ground truth. 
As can be observed, Particle-SfT provides the lowest value of 3D error in comparison to the other methods. This is more apparent in the datasets with lower number of matches.
In the last dataset, there are several discontinuities in the 3D error graph of state-of-the-art methods. 
This is due to the failure of shape inference in those images of the video by those methods.
Particle-SfT, however, succeeds to infer the 3D shape of the object in all of the images with a reasonable error. 


For the second test, we ran $\tt{ROBUSfT}$ on the public dataset provided in~\cite{varol2012monocular}. The dataset includes the 2D correspondences as well as 3D Kinect data of 193 consecutive images of a deforming paper. The paper is planar and no occlusion appears in the series of images. We compared our results with the results of Famouri et al.~\cite{famouri2018fast} and Ngo et al.~\cite{ngo2015template} which were presented in their papers. This is shown in Figure~\ref{image_3d_comparison}-a and Figure~\ref{image_3d_comparison}-b. As can be observed, $\tt{ROBUSfT}$ is both faster and more precise. It should be noted that $\tt{ROBUSfT}$ used directly images as the input and covered the whole process from extracting keypoints to 3D shape inference. In contrast, the other two algorithms used the already available correspondences in the dataset. Another relevant point is that in this test we use a serial CPU-GPU architecture instead of a parallel one. This is done to make sure that the captured image that we analyze and the ground truth that we compare to are for the same image. This consequently reduces the execution speed of our code compared to the parallel architecture. In the next series of tests we use the parallel architecture. 

\subsection{Evaluation of $\tt{ROBUSfT}$} \label{subsec:Eexperiments-with-real-data}
We first evaluate the efficiency of $\tt{ROBUSfT}$ in three real cases. These cases are shown in Figure~\ref{real_experiment}. The tested objects are a Spiderman poster, a chopping mat, and a t-shirt. In each case, the object is deformed in front of a 3D camera with which we capture both RGB image and the depth of each point on the object. We use the measured depth as ground truth for evaluating the reconstructed 3D shape. We use the Intel RealSense D435 depth camera and built-in libraries for aligning the depth map to the RGB image.  
For each case, four images of the experiment are shown in Figure~\ref{real_experiment}. 
In the first case, we set the resolution of the camera to $640\times480$. In the second and third cases, we increased it to $1280\times720$ due to the insufficient number of detected keypoints using the previous resolution. 
Below each image, the reconstructed 3D shape of the deforming object along with the 3D coordinates of the estimated correct matches (red particles) as well as their ground truth (green particles) are shown. The 3D coordinates of the estimated correct matches are estimated by calculating their barycentric coordinates in $\mathcal{P}$ with respect to $M$ and applying these coordinates on the 3D reconstructed mesh of the object. 
The number written below each frame is the median distance between the reconstructed 3D coordinates of the estimated correct matches and their ground truth. The median is chosen due to the probable existence of mismatches among the list of estimated correct matches. 
In 3D space, the ground truth of these mismatches can be located in the background and not on the object itself. 
This significantly increases the 3D shape error.
Using the median gives a better estimate of the 3D shape error considering the existence of this small percentage of mismatches with large 3D errors.

As can be observed, the pipeline succeeds to infer the 3D shape of the object in all of the cases.
This success is more visible in the second and third cases due to the relative scarcity of keypoints and existence of repetition in their patterns.
Regarding the Spiderman poster case, it should be noted that there are self-occlusions in the first and third illustrated images. In these images, the 3D shape of the object in the occluded areas is estimated by the deformation constraints implemented in Particle-SfT. These constraints preserve the geodesic distance between each pair of mesh points as its initial value in $M_T$. Regarding the runtime, using the parallel architecture and $640\times480$ captured frames as the input (as in the Spiderman poster case), the execution speed reaches 30\,fps.

The last experiment is a practical use case with robots. 
The experiment aims at highlighting the advantage of using known 3D coordinates in $\tt{ROBUSfT}$. As mentioned in \textit{Step 4} and shown in Figure~\ref{robot_experiment}, these known coordinates are an optional input to the last step of $\tt{ROBUSfT}$.
Their usage can increase the robustness of the tracking process. 
The setup of this experiment is the same as in~\cite{shetab2022rigid}, where we applied $\tt{ROBUSfT}$ in a robotic case, specifically, controlling the shape of deformable objects. 
The setup consists of two robotic arms grasping and manipulating the Spiderman poster from both sides and a top camera facing the manipulation area.
The 3D positions of the two robotic grippers are known in camera coordinates thanks to the known pose of each gripper in the robots' coordinate frames and also the external calibration between the robots and the camera.
For each gripper, we consider the closest mesh point to the gripper as a constrained mesh point.
These mesh points should be bound to their corresponding gripper and move with it. 
As described in~\cite{shetab2022rigid}, this binding is performed using a soft constraint. 
In this soft constraint, for each gripper, a sphere with a small radius centered at the gripper's 3D position is considered. Then, in each iteration of Particle-SfT, if the corresponding mesh point is outside this sphere, it will be absorbed to the closest point on the sphere surface. 
This soft constraint has two main advantages over rigidly binding the constrained mesh points to the grippers: first, they let the position-based dynamic equations in Particle-SfT that preserve the distances between the mesh points be applied on the constrained mesh points, which leads to a smoother reconstructed shape. Second, by applying soft constraints, we can cope with small possible errors in robot-camera calibration. In fact, a wrong robot-camera calibration leads to a wrong transfer of the grippers' 3D coordinates to the camera coordinate frame which eventually results in wrong coordinates of the constrained mesh points. 
By using the soft constraint and considering a sphere rather than a rigid bind, we give a certain degree of flexibility to the constrained mesh points to move in close proximity to the gripper's coordinates. This can compensate for slightly inaccurate coordinates of the grippers. 


\section{Conclusion} \label{sec:conclusion}
We have proposed $\tt{ROBUSfT}$, a new pipeline that can effectively reconstruct the 3D shape of an isometrically deforming object using a monocular 2D camera. 
The proposed pipeline addresses the well-known challenges in this area. These challenges include ambiguities in inferring the 3D shape of the deforming object from a single 2D image and real-time implementation.
We have introduced $\tt{myNeighbor}$, a novel mismatch removal algorithm for deforming objects, which works based on the preservation of the neighborhood structure of matches. We validated the efficiency of $\tt{myNeighbor}$ in comparison to the state-of-the-art algorithms in numerous experiments. 
In order to compare $\tt{ROBUSfT}$ and $\tt{myNeighbor}$ with the state-of-the-art methods in the literature, we have presented a novel type of experimental protocol called $\tt{FREX}$ (Fake Realistic Experiment). This protocol is executed once, but it provides a large number of resulting scenes of an isometrically deforming object in various conditions with 2D and 3D ground truth. This collection can be used to evaluate, compare, and validate algorithms regarding isometrically deforming objects. In addition, the provided 2D and 3D ground truth may be used for training learning-based algorithms.
In contrast to other artificially made scenes of an isometrically deforming surface, the generated images in our protocol are the result of real isometric deformations.

Possible directions for future work include \textit{(i)} exploiting the silhouette of the object in the image for improving 3D shape inference in challenging cases such as weakly-textured objects, \textit{(ii)} extending $\tt{ROBUSfT}$ to volumetric objects and \textit{(iii)} adding self-occlusion reasoning.

\section*{Acknowledgments}
This work was supported by project SOFTMANBOT, which received funding from the European Union’s Horizon 2020 research and innovation programme under grant agreement No 869855.
%
%
%
\balance

\bibliographystyle{ieeetr}
\bibliography{main}

\end{document}